\documentclass[12pt]{article}

\usepackage[a4paper, total={7in, 10in}]{geometry}

\usepackage[utf8]{inputenc}
\usepackage[T1]{fontenc}
\usepackage{helvet}

\usepackage{graphicx}
\usepackage{booktabs}
\usepackage{multirow}
\usepackage{threeparttable}
\usepackage{makecell}
\usepackage{array}
\usepackage{tabularx}
\usepackage{caption}
\usepackage{subcaption}
\usepackage{placeins}
\usepackage{siunitx}

\usepackage{amsmath}
\usepackage{amssymb}
\usepackage{amsfonts}
\usepackage{nicefrac}
\usepackage{pifont}

\usepackage{xcolor}
\usepackage{microtype}
\usepackage{url}
\usepackage{listings}

\usepackage{authblk}
\usepackage[title]{appendix}
\usepackage[numbers,sort&compress]{natbib}

\usepackage[hidelinks]{hyperref}
\usepackage{orcidlink}

\theadset{\bfseries}

\title{Improving Fairness of Large Language Model--Based ICU Mortality Prediction via Case-Based Prompting}

\author[1]{Gangxiong Zhang}
\author[1]{Yongchao Long\orcidlink{0009-0003-8345-2212}}
\author[1,2,*]{Yuxi Zhou}
\author[2]{Yong Zhang}
\author[3,4]{Shenda Hong\orcidlink{0000-0001-7521-5127}}

\affil[1]{School of Computer Science and Engineering, Tianjin University of Technology, Tianjin, China}
\affil[2]{DCST, BNRist, RIIT, Institute of Internet Industry, Tsinghua University, Beijing, China}
\affil[3]{National Institute of Health Data Science, Peking University, Beijing, China}
\affil[4]{Institute for Artificial Intelligence, Peking University, Beijing, China}
\affil[*]{Correspondence: joy\_yuxi@pku.edu.cn}

\date{}

\begin{document}

\maketitle

\section*{ABSTRACT}
Accurately predicting mortality risk in intensive care unit (ICU) patients is critical for clinical decision-making. Large language models (LLMs) are increasingly explored for clinical prediction using structured medical data, but their outputs may exhibit demographic disparities. Mitigating such disparities without degrading predictive performance remains challenging.

We systematically investigate demographic bias in LLM-based ICU mortality prediction and propose Case-Based Prompting (CAP), a training-free framework designed to improve the empirical balance between predictive performance and subgroup fairness. CAP retrieves clinically similar historical misprediction and demographic-sensitive cases with known outcomes and incorporates them as case-level contextual evidence. We further evaluate model discrimination, subgroup fairness, and feature-dependence consistency.

On MIMIC-IV, CAP improved AUROC from 0.806 to 0.873 and AUPRC from 0.497 to 0.694 compared with the Baseline Prompt. CAP also reduced several subgroup disparities, particularly for sex and White--Black comparisons, although age-related disparities remained non-negligible. Controlled demographic-information ablations showed that removing demographic information did not uniformly improve both predictive performance and subgroup fairness, indicating distinct performance--fairness patterns rather than a uniform benefit from demographic suppression. Feature-dependence analysis showed generally consistent expressed clinical factor patterns across demographic subgroups.

These findings suggest that CAP is a promising inference-time strategy for LLM-based clinical prediction under the evaluated retrospective setting. Further validation across additional LLMs, external datasets, and prospective clinical scenarios is required before deployment.

\section*{KEYWORDS}
Large language models; intensive care unit; mortality prediction; fairness; bias mitigation; case-based prompting; clinical decision support

\newpage
\section{INTRODUCTION}

In high-risk clinical environments, concerns regarding the fairness of artificial intelligence (AI) systems have become a central ethical focus in both academia and industry\cite{liu2025scoping}. As AI technologies become increasingly integrated into healthcare, their role in supporting clinical decision-making has grown rapidly\cite{zhang2024integrating,cummins2025hospital}. In intensive care units (ICUs), machine learning models have been widely explored for mortality risk prediction and illness severity assessment\cite{liu2023illness,hong2021predicting}. Traditional ICU mortality prediction models, such as APACHE, SAPS, SOFA-based risk scores, and machine learning models built on structured clinical variables, have demonstrated strong predictive performance in risk stratification. However, these models usually rely on predefined feature representations and fixed prediction pipelines, and they do not directly provide natural-language reasoning processes that can be inspected for potential demographic shortcuts or clinically unsupported explanations.

Large language models (LLMs) are increasingly being explored as general-purpose clinical reasoning interfaces. Although structured ICU mortality prediction remains a tabular-data task, LLMs provide a useful setting for studying fairness because they can transform structured clinical variables into natural-language risk assessments, generate explicit reasoning chains, and incorporate retrieved examples through in-context learning. Therefore, our goal is not to claim that LLMs are inherently superior to conventional tabular models for ICU mortality prediction. Rather, we focus on whether LLM-based clinical prediction systems exhibit demographic disparities when applied to structured ICU data, and whether such disparities can be mitigated through a lightweight inference-time intervention.

Fairness concerns are particularly important in ICU mortality prediction because systematic performance disparities associated with demographic attributes, such as sex, age, and race/ethnicity, may lead to inequitable allocation of monitoring intensity, clinical attention, or medical resources, thereby exacerbating existing health disparities\cite{meng2022interpretability,zack2024assessing,hasanzadeh2025bias}. For example, two ICU patients may have similar severity scores, vital signs, laboratory measurements, and treatment interventions, but differ in demographic attributes. In such cases, an LLM may produce different mortality estimates or explanations by overemphasizing age, sex, or race/ethnicity rather than objective physiological evidence. This type of demographic shortcut reasoning is especially concerning in high-stakes clinical settings, where even small differences in estimated mortality risk may influence clinical prioritization and trust in AI-assisted decision support. 

Demographic bias in LLM-based clinical prediction may arise from several sources. First, LLMs may inherit statistical associations, stereotypes, and uneven disease descriptions from pretraining corpora, including imbalanced representations of risk across sex, age, and race/ethnicity groups. Second, clinical notes and electronic health records may themselves reflect historical inequities in care delivery, documentation practices, treatment access, and subgroup representation. Third, when structured ICU variables are converted into natural-language prompts, demographic tokens may become salient contextual cues and may be over-weighted by the model, especially when clinical evidence is incomplete, noisy, or ambiguous. As a result, an LLM may rely on age, sex, or race/ethnicity as convenient proxies for risk rather than sufficiently integrating objective physiological evidence.

Given these potential sources of demographic shortcut reasoning, LLMs introduced into this high-risk clinical scenario face two key challenges. First, it is necessary to detect demographic bias across patient subgroups, including uneven performance for female, older, or racial/ethnic minority patients\cite{singhal2023large}. Conventional fairness metrics primarily measure output-level disparities, but they may fail to capture latent feature-dependence bias, namely the tendency of a model to rely on different clinical factors or reasoning patterns across demographic groups. Second, bias mitigation should be achieved without substantially sacrificing predictive performance. Existing mitigation strategies often rely on reweighting, fairness constraints, or model retraining, which may not be feasible when model parameters are inaccessible and may also degrade overall accuracy in high-stakes clinical settings.

Therefore, this study aims to systematically investigate the following three research questions:
\begin{itemize}
\item \textbf{Is bias present?} Do LLMs exhibit measurable performance disparities across demographic subgroups in ICU mortality prediction?
\item \textbf{Where is bias?} Are disparities confined to prediction outputs, or are they also reflected in feature reliance and reasoning patterns?
\item \textbf{Can bias be mitigated without indiscriminately suppressing demographic information?} Can a training-free prompting strategy improve the empirical balance between predictive performance and subgroup fairness while preserving potentially useful demographic context?
\end{itemize}

To address these questions, we construct a multidimensional bias assessment framework and propose \textbf{Case-Based Prompting (CAP)}, a training-free inference-time prompting framework for LLM-based ICU mortality prediction. CAP retrieves clinically similar historical misprediction and demographic-sensitive cases with known outcomes and incorporates them into the prompt as case-level contextual evidence. By providing concrete examples of prior prediction errors and known outcomes, CAP aims to guide the model toward more clinically grounded reasoning and reduce unsupported demographic shortcuts.

The main contributions of this work are as follows:
\begin{itemize}
\item \textbf{We develop a multidimensional bias evaluation framework} that integrates model discrimination, subgroup fairness, and feature-dependence analysis, enabling systematic assessment of both output-level disparities and reasoning-pattern differences across demographic groups.
\item \textbf{We propose Case-Based Prompting (CAP), a training-free prompting framework} that retrieves clinically similar historical misprediction and demographic-sensitive cases to provide case-level contextual evidence for LLM-based ICU mortality prediction.
\item \textbf{We conduct a retrospective evaluation on the real-world MIMIC-IV dataset} to examine whether CAP improves the empirical balance between predictive performance and subgroup fairness, while also analyzing feature-dependence consistency and representative case-level reasoning patterns.
\item \textbf{We conduct controlled demographic-information ablations} to separately evaluate demographic information presented to the LLM and demographic filtering used during retrieval, thereby empirically characterizing the predictive contribution of demographic context and the fairness--performance trade-off.
\end{itemize}

\section{Related Works}
\subsection{Bias and Fairness in Clinical AI}
Predictive models in healthcare, especially those deployed in high-risk settings such as intensive care units (ICUs), have been widely reported to exhibit disparities across demographic attributes including race, sex, and age \cite{shi2025large,zhang2025toward,wan2025evaluating}. Such disparities may arise from insufficient subgroup representation, historical inequities in healthcare delivery, and documentation bias embedded in electronic health records. A recent scoping review reported that a substantial proportion of clinical machine learning studies showed evidence of bias, often disproportionately affecting vulnerable patient groups \cite{colacci2025unequal}.

Fairness concerns are particularly important in ICU mortality prediction, where biased risk estimates may affect monitoring intensity, treatment prioritization, and resource allocation. For example, van Schaik et al. \cite{van2024monitoring} examined ICU mortality prediction models and showed that aggregate performance metrics alone are insufficient to reveal disparities across race, sex, and diagnostic subgroups. More broadly, fairness metrics such as demographic parity, equal opportunity, equalized odds, false-positive-rate gaps, false-negative-rate gaps, and subgroup discrimination gaps have been widely discussed in the machine learning fairness literature \cite{mehrabi2021survey}. Recent work further emphasizes that observed fairness gaps should be statistically validated, because subgroup differences may arise from sampling variability rather than systematic unfairness \cite{uddin2024novel,uddin2026fairnesspreserving}. These studies motivate our use of multiple fairness metrics together with statistical validation.

\subsection{LLMs for Clinical Prediction and Medical Fairness}
Large language models (LLMs) are increasingly being explored for healthcare tasks, including medical question answering, clinical reasoning, and prediction. Brown et al. \cite{brown2025large} evaluated GPT-3.5 and GPT-4 on clinical prediction tasks using VUMC and MIMIC-IV data and found that their predictive performance and calibration were generally inferior to locally trained machine learning models, while fairness patterns differed across demographic subgroups. This suggests that predictive performance and subgroup fairness may not improve simultaneously in LLM-based clinical prediction.

Beyond structured clinical prediction, recent medical AI studies have highlighted the importance of reasoning ability, multimodal evidence integration, disease-specific model development, and human-centered trustworthiness in healthcare applications. These include LLM reasoning in healthcare and medicine~\cite{lin2026surveyllmreasoning}, cross-modal knowledge diffusion for medical visual question answering~\cite{lin2025crossmodal}, specialized medical imaging models such as FATE-YOLO for brain tumor detection~\cite{yin2026fateyolo}, and human-centered trustworthy AI frameworks emphasizing explainability, human-in-the-loop design, hybrid AI, and uncertainty quantification~\cite{kohan2026humancentered}. These studies suggest that medical AI systems should be evaluated not only by predictive performance, but also by reasoning reliability, interpretability, and trustworthiness.

Recent studies have also begun to examine fairness and demographic bias in medical LLMs. Lu et al. \cite{lu2025fairmedicaladvice} investigated bias in LLM-based healthcare advice and explored mitigation strategies through fine-tuning and prompt engineering. FairMedQA provides an adversarial benchmark for evaluating demographic bias in medical question answering by constructing counterfactual question pairs across sensitive attributes \cite{xiao2025fairmedqa}. Similarly, counterfactual fairness alignment has been explored for medical LLMs, showing that demographic perturbation can help reveal unfair output patterns \cite{poulain2024aligningmedicalfairness,mahajan2025cognitive}. Compared with medical question answering or advice generation, ICU mortality prediction requires calibrated risk estimation from structured physiological variables. Therefore, fairness evaluation in this setting should consider not only counterfactual output changes, but also subgroup-level errors, false-negative rates, and feature-reliance patterns.

\subsection{Retrieval-Augmented and Case-Based Prompting}
Retrieval-augmented generation and case-based reasoning provide LLMs with external examples, documents, or cases that are relevant to the current query. Foundational work on few-shot prompting and in-context learning has shown that LLMs can adapt their behavior based on demonstrations provided in the prompt \cite{brown2020language}. Further studies on demonstration selection and nearest-neighbor in-context learning suggest that the choice of retrieved examples can substantially influence model behavior, particularly in few-shot or zero-shot settings \cite{wang2023large,ma2023fairness}. In clinical reasoning, case-based comparison is also intuitively aligned with how clinicians refer to similar historical cases when making decisions.

However, most retrieval-augmented prompting methods primarily aim to improve factuality, contextual relevance, or predictive accuracy, while their role in fairness-aware clinical prediction remains underexplored. For ICU mortality prediction, retrieved cases may be useful not only because they provide clinically similar examples, but also because historical misprediction and demographic-sensitive cases can provide corrective contextual evidence. This motivates our proposed \textbf{Case-Based Prompting (CAP)} framework, which retrieves clinically similar historical misprediction cases and demographic-sensitive cases with known outcomes to guide LLM reasoning during inference.

\subsection{Fairness-Aware Prompting and LLM Debiasing}
Bias mitigation methods can be broadly categorized into data-level, model-level, and inference-time interventions. Data- and model-level methods, such as reweighting, fairness constraints, adversarial learning, group-robust optimization, and distillation-based debiasing, often require access to training data or model parameters \cite{zhang2025towards}. For deployed LLMs, such access may be limited, making inference-time debiasing strategies more practical.

Prompt-based debiasing provides a lightweight alternative for regulating LLM behavior without updating model parameters. Prompting Fairness integrates causal reasoning into prompt design to reduce the influence of social information on LLM decisions \cite{liprompting}. Metric-Fair Prompting guides LLMs to make decisions under metric-fairness constraints \cite{wang2025metricfairprompting}. Self-Debias introduces a self-correction framework to interrupt bias propagation in LLM reasoning trajectories \cite{feng2026selfdebias}, while C2PO diagnoses and suppresses bias shortcuts through causal-contrastive preference optimization \cite{feng2025c2po}. These studies demonstrate the potential of inference-time interventions for reducing biased outputs.

Beyond prompt-level debiasing, recent studies have also explored training- or system-level fairness interventions in related AI tasks, such as Dual-Teacher Debiasing for mitigating domain bias through teacher--student distillation~\cite{li2023dualteacher} and FairCRS for improving user-oriented fairness in conversational recommendation systems~\cite{liu2024faircrs}. Although these methods address different task settings from structured ICU mortality prediction, they highlight the broader importance of diagnosing and mitigating bias in AI-driven decision systems.

Nevertheless, most existing LLM debiasing and fairness-intervention studies focus on general NLP, recommendation, or benchmark classification tasks. Their direct application to structured clinical prediction remains limited. Moreover, simply instructing an LLM not to rely on protected attributes may be insufficient in clinical settings, because demographic attributes can sometimes reflect clinically meaningful risk differences rather than unfair bias. CAP differs from generic fairness prompts by providing concrete, clinically similar historical cases with known outcomes, allowing the model to use case-level contextual evidence rather than abstract fairness instructions alone.

\section{METHODS}


\subsection{Problem Definition}

The objective of this study is to predict in-hospital mortality risk for ICU patients using structured clinical variables collected during the first 24 hours after ICU admission. Let
\[
\mathcal{D}=\{(\mathbf{x}_i,\mathbf{a}_i,y_i)\}_{i=1}^{n}
\]
denote the study cohort, where $\mathbf{x}_i \in \mathbb{R}^{d}$ represents the non-demographic structured clinical features of patient $i$, $\mathbf{a}_i$ denotes demographic attributes, and $y_i \in \{0,1\}$ is the ground-truth in-hospital mortality label. The complete structured input includes demographic characteristics, laboratory measurements, vital signs, clinical severity scores, and treatment interventions. In this formulation, demographic attributes are represented separately by $\mathbf{a}_i$, whereas $\mathbf{x}_i$ includes the remaining clinical variables used for risk estimation.

Given the structured patient information, an LLM-based prediction function $f_{\theta}(\cdot)$ generates an estimated mortality probability:
\[
\hat{p}_i = f_{\theta}(\mathbf{x}_i,\mathbf{a}_i) \in [0,1],
\]
where $\hat{p}_i$ denotes the predicted probability of in-hospital death. A binary prediction can be obtained as
\[
\hat{y}_i = \mathbb{I}(\hat{p}_i \geq \tau),
\]
where $\tau$ is the decision threshold. In addition to the mortality probability, the LLM is required to output a confidence score, key contributing clinical factors, and a brief reasoning summary in a predefined structured format. These outputs allow us to evaluate not only prediction performance, but also whether the model relies on different clinical factors across demographic subgroups.

To assess subgroup-level disparities, we consider commonly used demographic dimensions in clinical fairness evaluation, including sex, age group, and race/ethnicity:
\[
\mathcal{A}_{\mathrm{sex}}=\{\text{male},\text{female}\}, \quad
\mathcal{A}_{\mathrm{age}}=\{\text{18--59},\text{60+}\}, \quad
\mathcal{A}_{\mathrm{race}}=\{\text{White},\text{Black},\text{Asian},\text{Other}\}.
\]
For each demographic dimension, we define subgroup sets such as $S_{\mathrm{male}}$, $S_{\mathrm{female}}$, $S_{\mathrm{18-59}}$, $S_{\mathrm{60+}}$, $S_{\mathrm{White}}$, $S_{\mathrm{Black}}$, $S_{\mathrm{Asian}}$, and $S_{\mathrm{Other}}$. Our primary fairness analyses are conducted on these marginal demographic subgroups, while intersectional subgroup patterns are considered as supplementary analyses when sample size permits.

We evaluate each prompting strategy from three perspectives: predictive performance, subgroup fairness, and feature-dependence consistency. Predictive performance is assessed using discrimination, calibration, and classification metrics. Subgroup fairness is assessed by comparing error rates and discrimination performance across demographic groups. Feature-dependence consistency is evaluated by analyzing whether the key clinical factors generated by the LLM remain consistent across patient subgroups. This formulation allows us to examine both output-level disparities and potential reasoning-pattern differences in LLM-based ICU mortality prediction.

\subsection{Overview of Case-Based Prompting}\label{overview}
We propose \textbf{Case-Based Prompting (CAP)}, a training-free inference-time prompting framework for LLM-based ICU mortality prediction. The core idea of CAP is to provide the LLM with clinically similar historical cases as contextual evidence, so that the model can make predictions by comparing the current patient with prior cases that contain known outcomes and documented baseline errors or demographic-sensitivity annotations.

\begin{figure*}[!h]
    \centering
    \includegraphics[width=\textwidth]{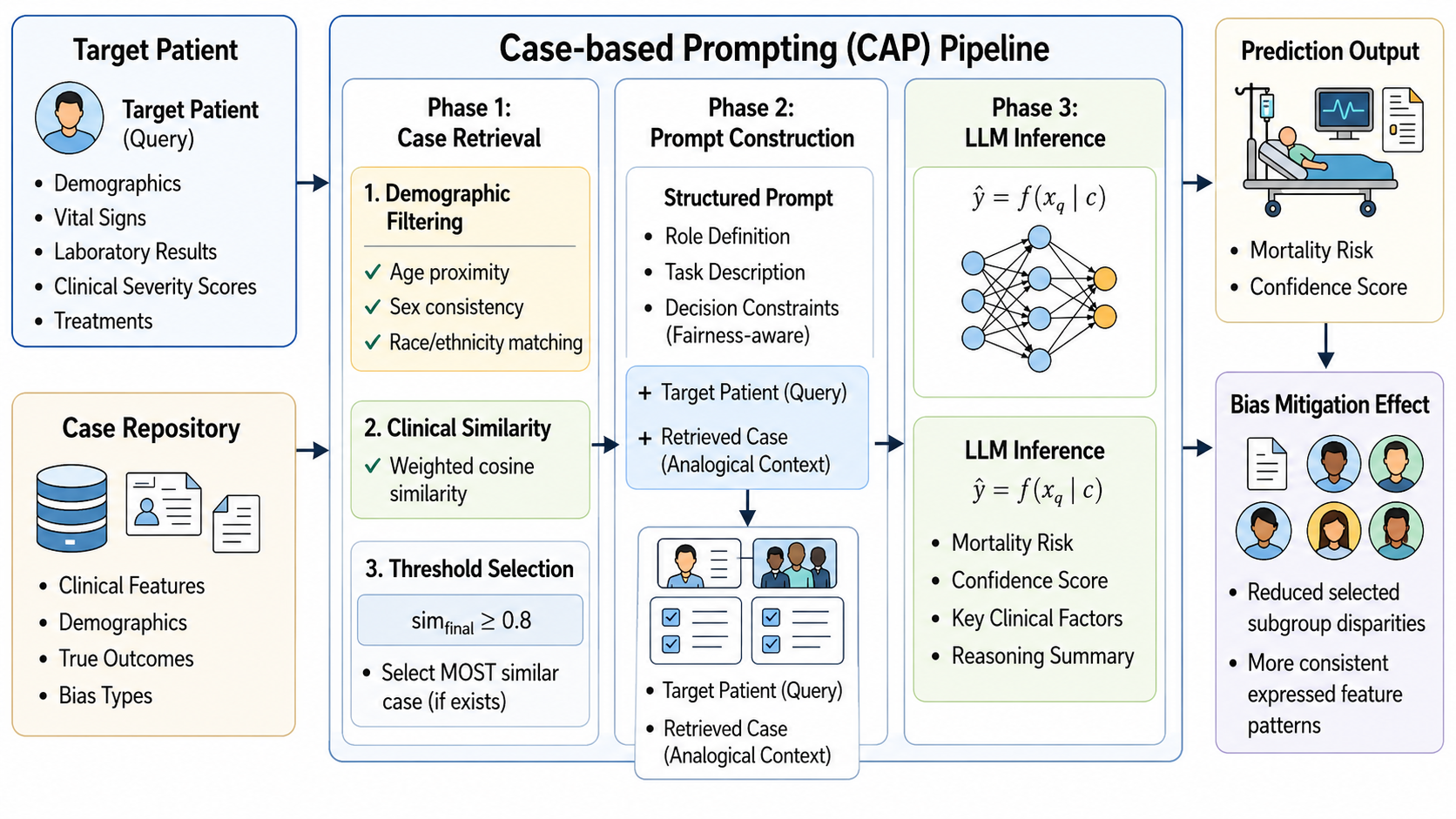}
    \caption{Overview of the Case-Based Prompting (CAP) framework for ICU mortality prediction. For a target patient, CAP retrieves clinically similar historical cases that were misclassified by the baseline prompting strategy and sensitive to counterfactual demographic perturbations from a curated case repository, and incorporates their structured features and known outcomes into the prompt. The framework is designed to provide case-level contextual evidence for LLM-based mortality prediction without model retraining.}
    \label{fig:framework}
\end{figure*}

As illustrated in Figure~\ref{fig:framework}, CAP consists of three main steps. First, for a target ICU patient, structured clinical variables from the first 24 hours after ICU admission are converted into a standardized patient summary. Second, CAP retrieves clinically similar historical cases from a curated case repository. The repository contains cases that are both misclassified by the baseline prompting strategy and sensitive to counterfactual demographic perturbations, and the retrieval procedure prioritizes clinical similarity after first using demographic attributes only to identify comparable candidate cases. Third, the retrieved case information, including patient features and known outcomes, is incorporated into the prompt together with the target patient information. The LLM then generates a mortality probability, a binary risk prediction, key contributing clinical factors, and a brief reasoning summary.

Different from generic fairness-aware prompts that only provide abstract instructions such as avoiding protected-attribute-based decisions, CAP introduces case-level contextual evidence into the reasoning process. This design is motivated by clinical decision-making practice, where physicians often compare a current patient with similar historical cases when assessing risk. By grounding the model's reasoning in clinically similar examples with known outcomes, CAP aims to reduce unsupported demographic shortcuts and encourage greater reliance on objective clinical evidence. Importantly, CAP does not require model retraining or parameter updates, making it applicable to settings where the underlying LLM is used as a frozen inference model.

The detailed construction of the case repository and the retrieval algorithm are described in the following subsections.

\subsection{Case Repository Construction}
\label{case_repository}

To support Case-Based Prompting, we constructed a repository of historical clinical cases from the training set only. The test set was not used during repository construction, retrieval hyperparameter selection, or case annotation, in order to avoid information leakage. The final repository was designed to include cases that satisfied two criteria: (i) they were misclassified by the baseline prompting strategy, and (ii) their predictions were sensitive to counterfactual demographic perturbations.

First, we applied the baseline LLM prompting strategy to the training set and identified misclassified samples by comparing the predicted binary label with the ground-truth in-hospital mortality label. False-positive (FP) cases were defined as patients incorrectly predicted to die during hospitalization, whereas false-negative (FN) cases were defined as patients incorrectly predicted to survive. Both error types were retained because they reflect clinically distinct failure modes: FP cases indicate risk overestimation, whereas FN cases correspond to missed high-risk patients. To reduce the influence of class imbalance, we applied a mild stratified sampling strategy to ensure that both FP and FN cases were represented in the repository.

Second, we performed counterfactual demographic perturbation on misclassified cases from the training set. For each candidate case, counterfactual variants were generated by modifying one demographic attribute at a time, such as sex, age group, or race/ethnicity, while keeping all non-demographic clinical variables unchanged. GPT-4 was used only as an auxiliary screening model to compare the outputs between the original and counterfactual versions of candidate cases. It was not used for final test-set prediction, model evaluation, ground-truth label generation, or direct mortality prediction. A case was considered demographic-sensitive if the predicted binary label changed after the perturbation or if the absolute change in predicted mortality probability exceeded 0.05. Formally, for a candidate case with original predicted probability $\hat{p}_i$ and counterfactual predicted probability $\hat{p}_i^{cf}$, the case was retained if
\begin{equation}
|\hat{p}_i - \hat{p}_i^{cf}| > 0.05
\end{equation}
or if the predicted binary label changed between the original and counterfactual versions.

From these screened cases, we selected 400 representative cases to construct the final repository. Therefore, all cases in the repository were both misclassified by the baseline prompting strategy and sensitive to at least one demographic perturbation. Each case was standardized into a structured format containing demographic attributes, vital signs, laboratory measurements, clinical severity scores, treatment interventions, baseline prediction, ground-truth outcome, error type, and demographic-sensitivity annotation. This standardized representation allowed retrieved cases to be consistently inserted into the prompt during inference.

To improve transparency and reproducibility, we report the composition of the case repository in Table~\ref{tab:case_repository}. The repository composition was fixed before final evaluation on the independent test set.

\begin{table}[t]
\centering
\caption{Composition of the case repository used by CAP. FP, false positive; FN, false negative.}
\label{tab:case_repository}
\begin{tabular}{lcc}
\toprule
Category & Subgroup / Type & Number (\%) \\
\midrule
\multirow{2}{*}{Prediction error} 
    & FP & 278 (69.5\%) \\
    & FN & 122 (30.5\%) \\
\midrule
\multirow{2}{*}{Sex} 
    & Male   & 227 (56.8\%) \\
    & Female & 173 (43.2\%) \\
\midrule
\multirow{2}{*}{Age group} 
    & 18--59 & 126 (31.5\%) \\
    & 60+    & 274 (68.5\%) \\
\midrule
\multirow{4}{*}{Race} 
    & White  & 217 (54.3\%) \\
    & Black  & 65 (16.2\%) \\
    & Asian  & 20 (5.0\%) \\
    & Other  & 98 (24.5\%) \\
\bottomrule
\end{tabular}
\end{table}

\subsection{Case Retrieval Algorithm}
\label{case_retrieval}

During inference, CAP retrieves historical cases that are clinically similar to the target patient and incorporates them into the prompt as contextual evidence. The retrieval algorithm was designed to prioritize clinical similarity while allowing demographic information to be used only as a fairness-auditing and representation-aware retrieval factor, rather than as direct evidence for mortality prediction.

For a query patient $q$ with structured clinical features $\mathbf{x}_q$, each candidate case $i$ in the repository was represented using the same normalized clinical feature space. Continuous variables were transformed using logarithmic transformation when appropriate and then range-normalized to reduce the influence of extreme values. Clinical similarity was computed using weighted cosine similarity:
\begin{equation}
\mathrm{sim}(\mathbf{x}_q,\mathbf{x}_i) =
\frac{(\mathbf{w} \odot \mathbf{x}_q) \cdot (\mathbf{w} \odot \mathbf{x}_i)}
{\| \mathbf{w} \odot \mathbf{x}_q \| \cdot \| \mathbf{w} \odot \mathbf{x}_i \|},
\end{equation}
where $\mathbf{w}$ denotes the normalized feature-importance weight vector and $\odot$ represents element-wise multiplication. The feature weights were derived from the normalized feature-importance scores of the XGBoost baseline trained on the training set. All feature weights were fixed before final evaluation on the independent test set.

To improve clinical plausibility, we further introduced a threshold-based penalty for physiologically important variables. If the query patient and a candidate case fell into different clinically meaningful intervals for key variables, such as severity scores, lactate, creatinine, oxygenation, or blood pressure, a penalty was applied. The exact threshold definitions were prespecified before test-set evaluation and are provided in the released implementation for reproducibility. The final similarity score was defined as:
\begin{equation}
\mathrm{sim}_{\mathrm{final}}(\mathbf{x}_q,\mathbf{x}_i) =
\mathrm{sim}(\mathbf{x}_q,\mathbf{x}_i) - \lambda N_{\mathrm{penalty}},
\end{equation}
where $N_{\mathrm{penalty}}$ is the number of violated clinical threshold conditions and $\lambda$ is the penalty coefficient, which was fixed at 0.1 before test-set evaluation. The penalty term was introduced to reduce the chance of retrieving cases that appear numerically similar but differ in clinically meaningful severity ranges.

Demographic attributes were retained in the structured prompt but were not designated as standalone evidence for mortality risk. During retrieval, age, sex, and race/ethnicity were used only in the first-stage candidate-filtering process to identify demographically comparable cases. Specifically, candidate cases were first restricted to those within an age window of $\pm5$ years of the query patient, with the same sex and race/ethnicity. Exact age was used internally only for the $\pm5$-year retrieval filter, whereas age was represented to the LLM using the predefined age-group categories of 18--59 and 60+. Among the filtered candidates, clinical similarity was then calculated using the weighted cosine similarity and clinical-threshold penalty defined above. Candidate cases with a final similarity score of $\mathrm{sim}_{\mathrm{final}} \geq 0.8$ were retained, and the most similar case was inserted into the CAP prompt when available. The threshold of 0.8 was selected based on preliminary retrieval coverage analysis on the training set, which showed that this value provided a reasonable balance between clinical similarity and retrieval availability. If no candidate case satisfied the threshold, CAP reverted to the corresponding non-case prompt without adding an analogy case. In our experiments, this fallback occurred in approximately 30\% of test cases, indicating that the retrieval criterion was relatively conservative. The influence of demographic information during retrieval was examined through retrieval-strategy ablations comparing Clinical-only Retrieval, Demographic-only Retrieval, Demographic-aware Retrieval, and Full CAP. To further isolate the effects of demographic information presented to the LLM and demographic filtering used during retrieval, controlled ablations were conducted as described in Section~\ref{demographic_ablation_methods}.

\subsection{Role of Demographic Attributes in CAP}
\label{demographic_attributes_cap}
CAP is not intended to be demographic-blind. Rather, it is designed to be demographic-aware but clinically grounded. Demographic attributes may contain clinically relevant contextual information in ICU mortality prediction. Age may be associated with physiological reserve, comorbidity burden, and baseline mortality risk, whereas sex may be relevant in specific disease or treatment contexts. Race/ethnicity requires more cautious interpretation because observed associations may also reflect social determinants of health, healthcare access, documentation practices, treatment patterns, and historical inequities rather than direct biological risk.

To avoid conflating fairness with demographic information removal, demographic attributes were retained in the structured patient representation and remained available in the target-patient input across the four primary prompting strategies. Importantly, we did not assume that demographic sensitivity itself constitutes unfair bias, nor did we attempt to infer causal clinical justification from retrospective observational data. Instead, demographic effects were interpreted at an empirical and operational level. A demographic effect was interpreted as potentially clinically contextualized when its influence on prediction co-occurred with relevant objective clinical evidence, such as severity scores, physiological instability, laboratory abnormalities, comorbidity burden, or treatment intensity. This operational criterion was used for auditing purposes and does not establish that the demographic effect was clinically or causally justified. Demographic subgroup differences or demographic terms appearing in model-generated reasoning were not, by themselves, considered evidence of either a clinically justified effect or unfair bias.

We separately defined \textit{demographic sensitivity} as a material change in model output after modifying one demographic attribute while holding all non-demographic clinical variables unchanged. A material change was defined as an absolute mortality-probability difference greater than 0.05 or a change in the binary prediction. Counterfactual demographic sensitivity alone was not considered sufficient evidence of an unsupported demographic shortcut. Instead, an unsupported shortcut was interpreted as demographic-dependent prediction or reasoning that was not adequately grounded in the available objective clinical context. This distinction was used as an operational auditing framework and does not establish a causal judgment about the demographic attribute.

Within CAP, demographic attributes have restricted and explicitly separated roles. Age, sex, and race/ethnicity may be used during first-stage retrieval filtering to identify broadly comparable candidate cases, with age restricted to a $\pm5$-year window around the query patient. Final case ranking is determined by weighted clinical similarity and clinical-threshold consistency rather than demographic similarity alone. Demographic information is also retained in the target-patient and retrieved-case descriptions for transparency and fairness auditing, but the prompt explicitly instructs the model not to treat demographic attributes as standalone evidence for mortality risk.

We empirically evaluated these design choices using controlled demographic-information ablation, counterfactual sensitivity analysis, and retrieval-strategy ablation. These analyses quantified the effects of retaining versus removing demographic information from the model input, characterized prediction changes under demographic perturbation while holding clinical variables constant, and isolated the contribution of demographic filtering during retrieval. Predictive performance and subgroup fairness were evaluated jointly to examine the predictive contribution of demographic context, counterfactual demographic sensitivity, potential shortcut behavior, and the fairness--performance trade-off.

\subsection{Prompting Strategies}
\label{prompting_strategies}
To evaluate how different prompting strategies affect both predictive performance and subgroup fairness, we designed four prompting strategies with increasing levels of guidance: Baseline Prompt, Fairness-Aware Prompt, System 2 Prompting, and Case-Based Prompting (CAP). All prompts followed a unified structure consisting of role definition, task description, patient information, decision constraints, and required output format. This unified design was used to ensure that differences across methods were primarily attributable to the prompting strategy rather than unrelated formatting variations.

The LLM was required to output a structured response containing the predicted in-hospital mortality probability, binary risk prediction, confidence score, key contributing clinical factors, and a brief reasoning summary. Complete prompt templates for all strategies are provided in Appendix~\ref{appendix:prompts}.

\begin{itemize}
\item \textbf{Baseline Prompt.}  
The baseline prompt includes only the basic instruction for ICU mortality prediction and the structured patient information. It does not include fairness-related instructions, step-by-step reasoning requirements, or retrieved case information.

\item \textbf{Fairness-Aware Prompt.}  
The fairness-aware prompt adds an explicit instruction that demographic attributes should not be used as standalone mortality-risk evidence. The model may consider demographic information when clinically relevant, but differences in predicted risk must be supported by the patient's overall clinical context and objective clinical evidence. This prompt serves as an instruction-based debiasing baseline.

\item \textbf{System 2 Prompting.}  
System 2 prompting combines fairness-aware instruction with structured step-by-step clinical reasoning. The model is guided to first evaluate physiological indicators, laboratory measurements, clinical severity scores, and treatment interventions, and then synthesize these clinical factors before generating the final mortality prediction. This strategy is intended to reduce intuitive or shortcut-based reasoning by encouraging a more explicit evidence-based reasoning process\cite{wei2022chain,furniturewala2024thinking,kamruzzaman2024prompting,jiao2024enhancing}.

\item \textbf{Case-Based Prompting (CAP).}  
CAP extends the structured reasoning prompt by incorporating retrieved historical cases from the case repository. Each retrieved case includes standardized clinical features, the baseline prediction, the ground-truth outcome, and demographic-sensitivity annotation. By providing clinically similar historical errors with known outcomes, CAP supplies case-level contextual evidence to support more clinically grounded risk estimation.
\end{itemize}

\begin{table}[htbp]
\centering
\caption{Comparison of prompting strategies for ICU mortality prediction.}
\label{tab:prompt_comparison}
\footnotesize
\renewcommand{\arraystretch}{1.08}
\setlength{\tabcolsep}{3.5pt}
\begin{tabular}{
p{2.4cm}
p{2.7cm}
p{4.0cm}
p{4.4cm}
}
\toprule
\textbf{Method} & \textbf{Core Design} & \textbf{Key Characteristics} & \textbf{Purpose} \\
\midrule

\textbf{Baseline Prompt}
& Task instruction
& Basic ICU mortality prediction without fairness instruction, structured reasoning, or retrieved cases
& Serves as the reference prompting strategy. \\

\midrule

\textbf{Fairness-Aware Prompt}
& Fairness instruction
& Adds an instruction to avoid unsupported reliance on demographic attributes
& Tests whether abstract fairness guidance can reduce subgroup disparities. \\

\midrule

\textbf{System 2 Prompting}
& Structured reasoning
& Guides step-by-step analysis of physiological indicators, laboratory values, severity scores, and interventions
& Encourages evidence-based reasoning and reduces shortcut reasoning. \\

\midrule

\textbf{Case-Based Prompting (CAP)}
& Retrieved case evidence
& Incorporates clinically similar historical misprediction and demographic-sensitive cases with known outcomes
& Provides case-level contextual evidence for clinically grounded prediction. \\

\bottomrule
\end{tabular}
\end{table}

\subsection{Controlled Demographic Information Ablation}
\label{demographic_ablation_methods}
To empirically examine how demographic information affects predictive performance and subgroup fairness, we conducted controlled ablation experiments that separately manipulated demographic information presented to the LLM and demographic information used during case retrieval. These analyses were designed to determine whether the observed fairness improvements of CAP could be explained by simply suppressing demographic information and to clarify the role of demographic filtering in the retrieval process.

In the \textit{Demographic-masked Prompt} setting, age group, sex, and race/ethnicity were removed from both the target-patient and retrieved-case descriptions, while the historical case retrieved by Full CAP was kept fixed. All non-demographic clinical variables, known outcomes, historical error information, model settings, output format, and parsing procedures were unchanged. Attribute-specific demographic-sensitivity annotations were removed from the prompt to prevent indirect disclosure of the masked demographic attribute. This setting therefore isolates the effect of demographic information directly presented to the LLM.

In the \textit{No-demographic Filtering} setting, demographic attributes remained available in the target-patient and retrieved-case descriptions, but age, sex, and race/ethnicity were excluded from first-stage candidate filtering. Candidate cases were instead ranked directly using the same weighted clinical similarity and clinical-threshold penalty as in Full CAP. All subsequent CAP components were kept unchanged. All ablation settings were evaluated on the same independent test set using the same Qwen3-32B backbone, inference parameters, decision threshold, and statistical procedures as the primary analysis. Predictive performance and subgroup fairness were evaluated using the metrics defined in Section~\ref{metrics_statistical_analysis}, and the corresponding results are reported in Section~\ref{demographic_information_ablation}.

\paragraph{Counterfactual demographic-sensitivity analysis.}
As a complementary patient-level audit of demographic dependence, we performed counterfactual demographic perturbation on the independent test set. For each evaluable patient, one demographic attribute was modified at a time while all non-demographic clinical variables were kept unchanged. Sex was switched between the male and female categories, age group was switched between 18--59 and 60+, and race/ethnicity was perturbed across the predefined racial/ethnic categories used in this study. For race/ethnicity, a prediction was considered race/ethnicity-sensitive if at least one valid counterfactual racial/ethnic category resulted in a material output change. For Full CAP, the identity and content of the historical case retrieved for the original patient were kept fixed during counterfactual perturbation. Therefore, this analysis was designed to assess sensitivity to demographic information presented to the LLM while avoiding changes introduced through the retrieval pathway. A prediction was considered demographic-sensitive if the absolute change in predicted mortality probability exceeded 0.05 or if the binary prediction changed. The reported sensitivity proportion was calculated as the number of demographic-sensitive predictions divided by the number of evaluable test patients with valid outputs under both the original and corresponding counterfactual conditions. This analysis was treated as a descriptive behavioral audit rather than a causal test of whether a demographic effect was clinically justified or unfair.

\subsection{Evaluation Metrics and Statistical Analysis}
\label{metrics_statistical_analysis}
To systematically evaluate LLM-based ICU mortality prediction, we assessed each prompting strategy from three complementary perspectives: predictive performance, subgroup fairness, and feature-dependence consistency. This design allowed us to examine both output-level disparities and potential differences in reasoning patterns across demographic groups.

\paragraph{Predictive performance.}
Overall predictive performance was evaluated using discrimination, classification, and calibration metrics. Discrimination was assessed by the area under the receiver operating characteristic curve (AUROC) and the area under the precision--recall curve (AUPRC). Classification performance was measured using F1-score, precision, sensitivity, specificity, false-positive rate (FPR), and false-negative rate (FNR). Probability calibration was assessed using the Brier score. All metrics were computed on the independent test set.

\paragraph{Subgroup fairness.}
Subgroup fairness was evaluated across demographic groups defined by sex, age group, and race/ethnicity. Given the clinical importance of missed high-risk patients in mortality prediction, we used the FNR gap as a primary fairness metric:
\begin{equation}
\Delta_{\mathrm{FNR}}(g,r) = \left| \mathrm{FNR}_{g} - \mathrm{FNR}_{r} \right|,
\end{equation}
where $g$ denotes the target subgroup and $r$ denotes the reference subgroup. We also reported the FPR gap:
\begin{equation}
\Delta_{\mathrm{FPR}}(g,r) = \left| \mathrm{FPR}_{g} - \mathrm{FPR}_{r} \right|,
\end{equation}
and the equalized odds gap:
\begin{equation}
\Delta_{\mathrm{EOdds}}(g,r) =
\max \left( \left| \mathrm{FNR}_{g} - \mathrm{FNR}_{r} \right|,
\left| \mathrm{FPR}_{g} - \mathrm{FPR}_{r} \right| \right).
\end{equation}
In addition, subgroup AUROC, AUPRC, and Brier score gaps were reported to characterize disparities in discrimination and calibration. Absolute subgroup gaps greater than 0.05 were interpreted as practically relevant differences and were considered together with confidence intervals and statistical tests, rather than as deterministic evidence of bias.

\paragraph{Feature-dependence consistency.}
To examine whether LLM reasoning patterns differed across demographic subgroups, we analyzed the three key clinical factors generated by the model for each patient. These factors were mapped to standardized clinical categories when necessary, such as severity scores, oxygenation, renal function, hemodynamic status, inflammatory markers, and treatment interventions. For each subgroup, we constructed a frequency distribution of the extracted key factors and compared subgroup-level feature-dependence patterns using Top-$K$ Jaccard similarity, cosine similarity, and Jensen--Shannon (JS) divergence. Higher Jaccard and cosine similarity, together with lower JS divergence, indicate more consistent feature-dependence patterns across subgroups. These metrics were interpreted as descriptive evidence of reasoning-pattern consistency rather than definitive proof of fairness.

\paragraph{Demographic-information ablation analysis.}
For the controlled demographic-information ablations, predictive performance and subgroup fairness were evaluated jointly using the same metrics as in the primary analysis. Full CAP was compared with variants that masked demographic information from the model input or removed demographic filtering from case retrieval. The mean FNR gap was calculated across the sex, age, White--Black, and White--Other comparisons; the White--Asian comparison was excluded from this aggregate measure because of the relatively small Asian subgroup. These comparisons were used to determine whether modifying the use of demographic information affected predictive performance, subgroup disparity, or both, thereby characterizing the empirical fairness--performance trade-off.

\paragraph{Statistical analysis.}
To avoid over-interpreting subgroup differences caused by random variation, 95\% confidence intervals for predictive performance and fairness metrics were estimated using patient-level bootstrap resampling with 1,000 iterations. For subgroup fairness gaps, permutation-based tests were performed by permuting demographic group labels while keeping predictions and ground-truth labels fixed. These permutation-test $p$-values assess whether the observed gap within each method is larger than expected under random group-label assignment, rather than directly testing the gap reduction between methods. Paired bootstrap resampling with 1,000 iterations was used for between-method comparisons because the evaluated methods were applied to the same test patients. For the controlled demographic-information analyses, Full CAP was compared with each ablation variant using paired differences in AUROC and mean FNR gap, with differences defined as the ablation value minus the Full CAP value. The 95\% confidence interval for each between-method difference was obtained from the 2.5th and 97.5th percentiles of the paired-bootstrap distribution. A between-method difference was considered statistically supported when its 95\% confidence interval excluded zero. Permutation-test $p$-values were used separately to assess whether within-method subgroup fairness gaps were larger than expected under random group-label assignment.

\subsection{Experimental Setup}
\label{experimental_setup}

All main LLM experiments were conducted using Qwen3-32B as the backbone model. The model parameters were kept frozen throughout all experiments, and no fine-tuning or parameter updates were performed. We evaluated four prompting strategies: Baseline Prompt, Fairness-Aware Prompt, System 2 Prompting, and the proposed Case-Based Prompting (CAP). All strategies used the same patient input format and structured output format to ensure comparability. Supplementary evaluations using Qwen3-8B and BioMistral-7B were conducted as lightweight-model robustness checks and are reported in Appendix~\ref{appendix:additional_llms}.

The LLM was required to output a structured JSON response containing the predicted in-hospital mortality probability, binary prediction, confidence score, key contributing clinical factors, and a brief reasoning summary. The predicted probability was extracted from the JSON field \texttt{probability}, and the binary prediction was derived using a threshold of 0.50. Outputs that did not conform to the predefined JSON format were re-queried once; if the second response remained invalid, rule-based extraction was applied. The same parsing rules were used for all prompting strategies.

To provide a conventional structured-data reference, we trained an XGBoost model on the same training set using the same structured clinical variables. XGBoost was included as a traditional tabular machine-learning baseline rather than as a direct LLM debiasing method.

The main evaluation was performed on an independent test set. Predictive performance was assessed using AUROC, AUPRC, F1-score, precision, sensitivity, specificity, and Brier score. Subgroup fairness and feature-dependence consistency were evaluated as described in Section~\ref{metrics_statistical_analysis}. Given the computational cost of LLM inference, each prompting strategy was evaluated once under fixed inference settings, and statistical uncertainty was estimated using patient-level bootstrap resampling.

All data preprocessing, case repository construction, case retrieval, model inference, output parsing, and statistical analyses were implemented in Python 3.10. Experiments were run on a workstation equipped with two NVIDIA RTX A6000 GPUs. The main inference settings are summarized in Table~\ref{tab:llm_settings}.

\begin{table}[t]
\centering
\caption{Main LLM inference settings.}
\label{tab:llm_settings}
\begin{tabular}{lc}
\toprule
Parameter & Value \\
\midrule
Backbone LLM & Qwen3-32B \\
Model update & Frozen; no fine-tuning \\
Temperature & 0.1 \\
Top-p & 0.9 \\
Maximum output tokens & 1024 \\
Random seed & 42 \\
Repeated runs & 1 \\
Output format & Structured JSON \\
Binary threshold & 0.50 \\
\bottomrule
\end{tabular}
\end{table}

\subsection{Ethics Statement}
This study used de-identified data from the MIMIC-IV database, which are available through PhysioNet subject to credentialed access. The collection of patient information and creation of the MIMIC-IV research resource were reviewed by the Institutional Review Board of Beth Israel Deaconess Medical Center, which granted a waiver of informed consent and approved the data-sharing initiative. The authors completed the required training and signed the data use agreement prior to accessing the data. All methods were performed in accordance with relevant guidelines and regulations.

\section{RESULTS}

\subsection{Study Population}
\label{study_population}
This study used the MIMIC-IV database to construct a retrospective ICU cohort. The inclusion criteria were adult patients who were admitted to the ICU for the first time and had a hospital stay longer than 24 hours. Before model development, data quality assessment was performed, and patients with excessively high missing-data ratios were excluded.

The final cohort included 45,153 patients. Patients were randomly divided into a training set and an independent test set at a 7:3 ratio. Missing values in retained variables were imputed using Multiple Imputation by Chained Equations (MICE). To avoid information leakage, imputation models and preprocessing parameters were fitted using the training set only and then applied to the independent test set. The training set was used for XGBoost model training, case repository construction, and preliminary retrieval-coverage analysis. The independent test set was used for all final evaluations, including predictive performance, subgroup fairness, feature-dependence consistency, and case-level analyses. The baseline characteristics of the study cohort are summarized in Table~\ref{tab:demo}. The training and test sets were broadly similar in their demographic, clinical, treatment, and outcome distributions. Given the large sample size, statistically significant differences were observed for selected variables despite relatively small absolute differences.

\begin{table}[htbp]
\centering
\caption{Baseline characteristics of the study cohort. SD, standard deviation. P-values were calculated using the t-test for continuous variables and the chi-square test for categorical variables to compare the training and test sets.}
\label{tab:demo}
\resizebox{\textwidth}{!}{
\begin{tabular}{lcccc}
\toprule
\textbf{Feature} & \textbf{Overall (n=45,153)} & \textbf{Train (n=31,607)} & \textbf{Test (n=13,546)} & \textbf{p-value} \\
\midrule

\multicolumn{5}{l}{\textbf{Demographics}} \\
Age (years), mean $\pm$ SD & 66.1 $\pm$ 16.2 & 66.1 $\pm$ 16.2 & 66.0 $\pm$ 16.2 & 0.55 \\
Female, n (\%) & 20,138 (44.6) & 14,072 (44.5) & 6,066 (44.8) & 0.62 \\
Race, n (\%) & & & & 0.45 \\
\quad White & 24,404 (54.1) & 17,107 (54.1) & 7,297 (53.9) &  \\
\quad Black & 7,043 (15.6) & 4,970 (15.7) & 2,073 (15.3) &  \\
\quad Asian & 512 (1.1) & 358 (1.1) & 154 (1.1) &  \\
\quad Other & 13,194 (29.2) & 9,172 (29.0) & 4,022 (29.7) &  \\[3pt]

\multicolumn{5}{l}{\textbf{Clinical Severity, mean $\pm$ SD}} \\
GCS score & 8.0 $\pm$ 4.6 & 8.0 $\pm$ 4.6 & 8.0 $\pm$ 4.6 & 0.35 \\
APACHE III score & 29.3 $\pm$ 19.5 & 29.3 $\pm$ 19.6 & 29.2 $\pm$ 19.4 & 0.41 \\
SOFA score (24 h) & 3.9 $\pm$ 2.4 & 3.9 $\pm$ 2.4 & 3.9 $\pm$ 2.4 & 0.48 \\
Charlson Comorbidity Index & 5.1 $\pm$ 3.0 & 5.1 $\pm$ 3.0 & 5.1 $\pm$ 3.0 & 0.52 \\[3pt]

\multicolumn{5}{l}{\textbf{Vital Signs (first 24 hours), mean $\pm$ SD}} \\
SpO\textsubscript{2,min} (\%) & 87.0 $\pm$ 4.4 & 87.0 $\pm$ 4.4 & 87.0 $\pm$ 4.4 & 0.65 \\
Heart rate (beats/min) & 87.8 $\pm$ 17.4 & 87.8 $\pm$ 17.5 & 87.7 $\pm$ 17.1 & 0.52 \\
Respiratory rate (breaths/min) & 19.3 $\pm$ 4.8 & 19.3 $\pm$ 4.8 & 19.2 $\pm$ 4.7 & 0.44 \\
Mean arterial pressure (mmHg) & 84.8 $\pm$ 13.2 & 84.8 $\pm$ 13.3 & 84.7 $\pm$ 13.1 & 0.47 \\[3pt]

\multicolumn{5}{l}{\textbf{Laboratory Measurements, mean $\pm$ SD}} \\
Creatinine\textsubscript{max} (mg/dL) & 1.6 $\pm$ 1.3 & 1.6 $\pm$ 1.3 & 1.6 $\pm$ 1.3 & 0.71 \\
Lactate\textsubscript{max} (mmol/L) & 2.1 $\pm$ 1.8 & 2.1 $\pm$ 1.8 & 2.1 $\pm$ 1.8 & 0.58 \\
Troponin\textsubscript{max} (ng/mL) & 0.6 $\pm$ 1.5 & 0.6 $\pm$ 1.5 & 0.6 $\pm$ 1.5 & 0.66 \\
Platelet\textsubscript{min} (10\textsuperscript{3}/$\mu$L) & 185.4 $\pm$ 98.6 & 185.6 $\pm$ 98.2 & 185.0 $\pm$ 99.3 & 0.59 \\
Bilirubin\textsubscript{max} (mg/dL) & 1.5 $\pm$ 2.1 & 1.5 $\pm$ 2.0 & 1.5 $\pm$ 2.2 & 0.63 \\
WBC\textsubscript{max} (10\textsuperscript{3}/$\mu$L) & 13.8 $\pm$ 7.2 & 13.8 $\pm$ 7.1 & 13.7 $\pm$ 7.3 & 0.55 \\
24-hour urine output (mL) & 1680.8 $\pm$ 1223.2 & 1651.4 $\pm$ 1202.4 & 1749.4 $\pm$ 1267.8 & <0.001 \\[3pt]

\multicolumn{5}{l}{\textbf{Treatment Interventions}} \\
Mechanical ventilation, n (\%) & 17,210 (38.1) & 12,039 (38.1) & 5,171 (38.2) & 0.87 \\
Code status (DNR), n (\%) & 3,840 (8.5) & 2,688 (8.5) & 1,152 (8.5) & 1.00 \\[3pt]

\multicolumn{5}{l}{\textbf{Outcomes}} \\
In-hospital mortality, n (\%) & 6,500 (14.4) & 4,535 (14.4) & 1,965 (14.5) & 0.67 \\
\bottomrule
\end{tabular}
}
\end{table}

\subsection{Predictive Performance Comparison}
\label{predictive_performance}

\begin{table}[htbp]
  \centering
  \small
  \caption{Performance comparison of LLM prompting strategies and the XGBoost tabular reference baseline. Metrics are reported with 95\% confidence intervals estimated using 1,000 patient-level bootstrap resamples. Statistical significance was evaluated against the Baseline Prompt among LLM-based methods using paired bootstrap tests (* $p<0.05$, ** $p<0.01$). Bold values indicate the best value for each metric. Lower Brier scores indicate better calibration.}
  \label{tab:qwen_xgboost_comparison}
  \setlength{\tabcolsep}{3pt}
  \begin{tabular}{llccccccc}
    \toprule
    \textbf{Model} & \textbf{Method} & \textbf{AUROC} & \textbf{AUPRC} & \textbf{F1} & \textbf{Prec} & \textbf{Sen} & \textbf{Spec} & \textbf{Brier} \\
    \midrule
    \multirow{8}{*}{\shortstack{Qwen3 \\ \scriptsize 32B}}

    & Baseline
    & 0.806 & 0.497 & 0.438 & 0.313 & 0.725 & 0.731 & 0.110 \\
    &
    & {\scriptsize (0.792--0.820)}
    & {\scriptsize (0.470--0.525)}
    & {\scriptsize (0.412--0.465)}
    & {\scriptsize (0.285--0.342)}
    & {\scriptsize (0.695--0.755)}
    & {\scriptsize (0.715--0.747)}
    & {\scriptsize (0.104--0.115)} \\

    & Fairness-Aware
    & 0.791 & 0.479 & 0.402 & 0.281 & 0.705 & 0.694 & 0.111 \\
    &
    & {\scriptsize (0.777--0.805)}
    & {\scriptsize (0.452--0.506)}
    & {\scriptsize (0.378--0.428)}
    & {\scriptsize (0.255--0.308)}
    & {\scriptsize (0.675--0.735)}
    & {\scriptsize (0.678--0.710)}
    & {\scriptsize (0.106--0.116)} \\

    & System 2
    & 0.785 & 0.475 & 0.426 & 0.322 & 0.632 & 0.774 & 0.110 \\
    &
    & {\scriptsize (0.771--0.799)}
    & {\scriptsize (0.448--0.502)}
    & {\scriptsize (0.402--0.452)}
    & {\scriptsize (0.294--0.350)}
    & {\scriptsize (0.600--0.664)}
    & {\scriptsize (0.760--0.788)}
    & {\scriptsize (0.105--0.116)} \\

    & CAP
    & \textbf{0.873}$^{**}$
    & \textbf{0.694}$^{**}$
    & \textbf{0.525}$^{**}$
    & \textbf{0.380}$^{*}$
    & \textbf{0.850}$^{**}$
    & 0.764
    & 0.100$^{**}$ \\
    &
    & {\scriptsize (0.860--0.886)}
    & {\scriptsize (0.665--0.723)}
    & {\scriptsize (0.500--0.551)}
    & {\scriptsize (0.350--0.410)}
    & {\scriptsize (0.825--0.875)}
    & {\scriptsize (0.749--0.779)}
    & {\scriptsize (0.095--0.106)} \\

    \midrule
    \multirow{2}{*}{XGBoost}
    & Tabular reference
    & 0.866 & 0.548 & 0.474 & 0.327 & 0.724 & \textbf{0.849} & \textbf{0.084} \\
    &
    & {\scriptsize (0.853--0.879)}
    & {\scriptsize (0.520--0.576)}
    & {\scriptsize (0.450--0.499)}
    & {\scriptsize (0.300--0.354)}
    & {\scriptsize (0.694--0.754)}
    & {\scriptsize (0.835--0.863)}
    & {\scriptsize (0.078--0.089)} \\

    \bottomrule
  \end{tabular}
\end{table}

Table~\ref{tab:qwen_xgboost_comparison} summarizes the predictive performance of different LLM prompting strategies and the XGBoost tabular reference baseline. Among the LLM-based methods, Case-Based Prompting (CAP) achieved the highest AUROC, AUPRC, F1-score, precision, and sensitivity. Compared with the Baseline Prompt, CAP improved AUROC from 0.806 to 0.873, AUPRC from 0.497 to 0.694, and F1-score from 0.438 to 0.525, corresponding to absolute increases of 0.067, 0.197, and 0.087, respectively. CAP also achieved the highest sensitivity among all evaluated methods, suggesting improved identification of patients who experienced in-hospital mortality.

The Fairness-Aware Prompt and System 2 Prompting did not improve overall discrimination compared with the Baseline Prompt. This suggests that abstract fairness instructions or step-by-step reasoning alone may be insufficient for improving predictive performance in this structured ICU mortality prediction task. In contrast, CAP provided retrieved historical cases with known outcomes, which may have offered additional contextual evidence for risk estimation.

XGBoost was included as a conventional tabular machine-learning reference rather than as a direct LLM debiasing baseline. XGBoost achieved the lowest Brier score and the highest specificity, indicating stronger calibration and better identification of lower-risk survivors. CAP showed higher discrimination and sensitivity in this evaluation, whereas XGBoost retained advantages in calibration and specificity. These results should therefore be interpreted as evidence that case-based prompting improves LLM-based prediction relative to other prompting strategies, while the XGBoost comparison provides context relative to a conventional structured-data model. Taken together, these findings suggest that CAP did not improve subgroup fairness by simply sacrificing overall predictive performance. Instead, under the evaluated setting, CAP improved discrimination and sensitivity while also reducing several subgroup disparities, as further examined in the following fairness and ablation analyses.

\subsection{Subgroup Fairness Evaluation with Statistical Validation}
\label{subgroup_fairness}
To evaluate subgroup-level fairness, we examined model performance across demographic dimensions, including sex, age group, and race/ethnicity. In addition to subgroup-specific AUROC, we focused on false-negative-rate (FNR) gaps because missed high-risk patients are clinically important in ICU mortality prediction. In this setting, the FNR gap corresponds to Equal Opportunity Difference (EOD), defined as the absolute difference in FNR between a target subgroup and a reference subgroup.

Following prior clinical bias-mitigation work, we used an absolute FNR gap of 0.05, corresponding to five percentage points, as a practical reference threshold for potentially meaningful subgroup disparity~\cite{mackin2025identifying}. This threshold was used for interpretability and comparison with prior work, but it was not treated as a statistical significance criterion. Therefore, subgroup fairness gaps were interpreted together with 95\% bootstrap confidence intervals and permutation-based statistical tests.

\begin{figure}[h!]
    \centering
    \includegraphics[width=\textwidth]{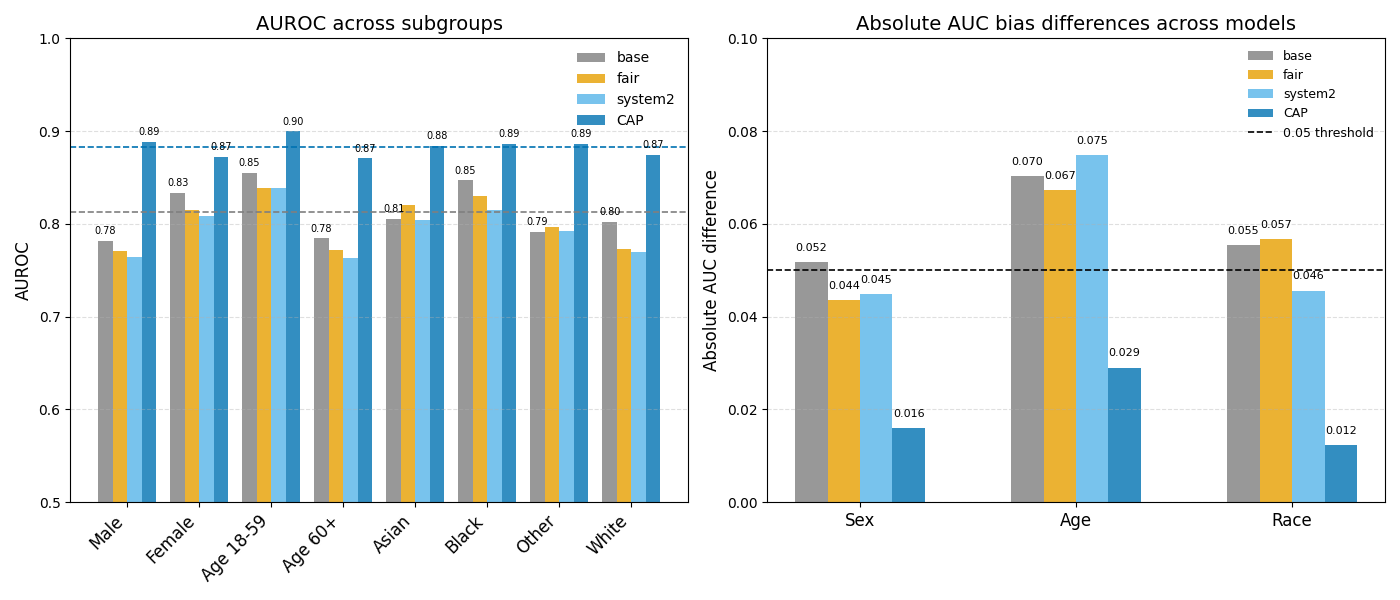}
    \caption{Subgroup-specific AUROC performance and disparity analysis across prompting strategies. Left panel: AUROC values across sex, age, and race/ethnicity subgroups. The horizontal dashed lines indicate the overall mean AUROC for the Baseline Prompt and CAP. Right panel: maximum absolute subgroup AUROC difference within each demographic dimension for each prompting strategy. The dashed line at 0.05 is shown as a practical visual reference for subgroup AUROC differences and should not be interpreted as a statistical significance threshold.}
    \label{fig:discrimination_bias}
\end{figure}

Figure~\ref{fig:discrimination_bias} shows subgroup-specific AUROC patterns across prompting strategies. The Baseline Prompt exhibited visible AUROC variation across demographic subgroups, with the largest disparity observed across age groups. Fairness-Aware Prompting reduced some subgroup differences but did not improve overall discrimination. CAP maintained high overall AUROC while reducing the maximum subgroup AUROC difference to approximately 0.03. These findings suggest that CAP showed smaller subgroup AUROC variation across several demographic dimensions, although subgroup AUROC differences alone are insufficient to fully characterize fairness.

Table~\ref{tab:fnr_gap_validation} reports FNR gaps across demographic comparisons. Compared with the Baseline Prompt, CAP reduced the sex-related FNR gap from 0.083 to 0.003 and the White--Black FNR gap from 0.058 to 0.005. CAP also reduced the White--Other FNR gap from 0.084 to 0.022. However, age-related disparity remained non-negligible, with the FNR gap decreasing from 0.180 to 0.126. These results indicate that CAP reduced several subgroup disparities, particularly for sex and selected race/ethnicity comparisons, but did not eliminate demographic disparities across all dimensions.

\begin{table}[h!]
\centering
\small
\caption{False-negative-rate (FNR) gaps across demographic subgroups. FNR gap is computed as the absolute difference in FNR between the target subgroup and the reference subgroup, corresponding to Equal Opportunity Difference (EOD) in this setting. Following prior clinical bias-mitigation work, 0.05 was used as a practical reference threshold for potentially meaningful subgroup disparity. Values are interpreted together with 95\% bootstrap confidence intervals and permutation-test $p$-values. The permutation-test $p$-values assess whether the observed subgroup FNR gap within each method is larger than expected under random demographic group-label assignment, rather than directly testing the gap reduction between methods. Lower values indicate smaller subgroup disparity.}

\label{tab:fnr_gap_validation}
\setlength{\tabcolsep}{3pt}
\begin{tabular}{llcccc}
\toprule
\textbf{Dimension} & \textbf{Comparison} & \textbf{Baseline} & \textbf{Fairness-Aware} & \textbf{System 2} & \textbf{CAP} \\
\midrule

\multirow{4}{*}{Sex} 
& Male vs Female 
& 0.083 
& 0.070 
& 0.035 
& \textbf{0.003} \\
& FNR 
& 0.306/0.223 
& 0.336/0.267 
& 0.242/0.207 
& 0.259/0.256 \\
& 95\% CI 
& (0.041--0.126) 
& (0.030--0.112) 
& (0.006--0.073) 
& (0.000--0.036) \\
& $p$-value 
& 0.018 
& 0.031 
& 0.032 
& 0.912 \\
\midrule

\multirow{4}{*}{Age} 
& 18--59 vs 60+ 
& 0.180 
& 0.164 
& 0.152 
& \textbf{0.126} \\
& FNR 
& 0.143/0.323 
& 0.191/0.355 
& 0.119/0.272 
& 0.170/0.296 \\
& 95\% CI 
& (0.129--0.232) 
& (0.112--0.218) 
& (0.101--0.204) 
& (0.075--0.179) \\
& $p$-value 
& $<0.001$ 
& $<0.001$ 
& $<0.001$ 
& 0.004 \\
\midrule

\multirow{12}{*}{Race/Ethnicity} 
& White vs Black 
& 0.058 
& 0.074 
& 0.046 
& \textbf{0.005} \\
& FNR 
& 0.251/0.193 
& 0.309/0.235 
& 0.221/0.175 
& 0.253/0.248 \\
& 95\% CI 
& (0.006--0.112) 
& (0.019--0.131) 
& (0.000--0.096) 
& (0.000--0.048) \\
& $p$-value 
& 0.046 
& 0.021 
& 0.071 
& 0.864 \\
\cmidrule{2-6}

& White vs Other 
& 0.084 
& 0.023 
& 0.036 
& \textbf{0.022} \\
& FNR 
& 0.251/0.336 
& 0.309/0.332 
& 0.221/0.257 
& 0.253/0.275 \\
& 95\% CI 
& (0.035--0.135) 
& (0.000--0.070) 
& (0.000--0.085) 
& (0.000--0.067) \\
& $p$-value 
& 0.012 
& 0.381 
& 0.174 
& 0.426 \\
\cmidrule{2-6}

& White vs Asian 
& 0.008 
& 0.031 
& \textbf{0.004} 
& 0.031 \\
& FNR 
& 0.251/0.243 
& 0.309/0.278 
& 0.221/0.217 
& 0.253/0.222 \\
& 95\% CI 
& (0.000--0.090) 
& (0.000--0.126) 
& (0.000--0.082) 
& (0.000--0.119) \\
& $p$-value 
& 0.842 
& 0.486 
& 0.936 
& 0.512 \\
\bottomrule
\end{tabular}
\end{table}

Overall, CAP showed the most consistent reduction in FNR gaps for sex and selected race/ethnicity comparisons. Under CAP, the sex-related FNR gap decreased below the practical 0.05 reference threshold, and its within-method permutation-test $p$-value was greater than 0.05. The White--Black and White--Other gaps were also reduced to below the practical threshold. In contrast, the age-related FNR gap remained above 0.05 after CAP and remained statistically significant, suggesting that age-related subgroup disparity was more difficult to reduce. The White--Asian comparison should be interpreted cautiously because the Asian subgroup was relatively small in the test set, which may lead to greater statistical uncertainty. The persistent age-related FNR gap under CAP also indicates that the method did not simply force demographic parity across all groups, but reduced several subgroup disparities while leaving more complex age-related differences only partially mitigated. Taken together, these findings support a cautious interpretation: CAP reduced several subgroup disparities under the evaluated setting, but demographic disparities were not fully eliminated. Additional fairness-related metrics, including FPR gap, equalized odds gap, subgroup discrimination gap, and calibration gap, are provided in Appendix~\ref{appendix:additional_fairness}.
\FloatBarrier

\subsection{Demographic-Use Retrieval Ablation}
\label{retrieval_ablation}
To characterize how different case-selection strategies affect predictive performance and subgroup fairness, we conducted a retrieval-strategy ablation analysis. Five strategies were compared: (i) Baseline Prompt without retrieval; (ii) Clinical-only Retrieval, which selected cases using normalized clinical variables only; (iii) Demographic-only Retrieval, which selected cases using age group, sex, and race/ethnicity only; (iv) Demographic-aware Retrieval, which applied demographic candidate filtering before clinical similarity ranking; and (v) Full CAP, which combined demographic-aware retrieval with the complete CAP case context, including historical error type and demographic-sensitivity annotation.

For Demographic-only Retrieval, each repository case received a demographic matching score equal to the number of attributes shared with the query patient across age group, sex, and race/ethnicity. The case with the highest score was selected, with ties resolved by random selection using a fixed seed of 42. One case was retrieved for each query patient. No clinical variables, feature-importance weights, clinical-similarity scores, or threshold penalties were used in this strategy.

In all retrieval-based strategies, the selected case's clinical features and known outcome were provided as auxiliary contextual evidence. Full CAP additionally included historical error type and demographic-sensitivity annotation. The term ``Demographic-only'' refers only to the criterion used for case selection and does not imply that mortality prediction itself was based only on demographic information. This strategy was included as a diagnostic comparison and was not part of the proposed CAP framework. In Full CAP, demographic attributes were used only during first-stage candidate filtering, whereas final case ranking was determined by weighted clinical similarity and clinical-threshold consistency. White--Asian comparisons were reported separately in the subgroup fairness analysis but were excluded from the aggregated race gap because of the relatively small Asian subgroup size. The results are summarized in Table~\ref{tab:retrieval_ablation}.

\begin{table}[htbp]
\centering
\small
\caption{Retrieval-strategy ablation for Case-Based Prompting. Clinical-only Retrieval uses clinical variables only for case selection, whereas Demographic-only Retrieval uses age group, sex, and race/ethnicity only. Demographic-aware Retrieval applies demographic candidate filtering before clinical similarity ranking. All retrieval-based strategies provide the selected case's clinical features and known outcome as auxiliary context, while Full CAP additionally includes historical error type and demographic-sensitivity annotation. Mean FNR gap is calculated across the sex, age, White--Black, and White--Other comparisons; race gap is the average of the White--Black and White--Other FNR gaps. Lower gap values indicate smaller subgroup disparity. Sen, sensitivity.}
\label{tab:retrieval_ablation}
\setlength{\tabcolsep}{3.5pt}
\begin{tabular}{lccccccc}
\toprule
\textbf{Strategy} & \textbf{AUROC} & \textbf{AUPRC} & \textbf{Sen} & \textbf{Mean FNR gap} & \textbf{Sex gap} & \textbf{Age gap} & \textbf{Race gap} \\
\midrule
Baseline Prompt 
& 0.806 & 0.497 & 0.725 & 0.101 & 0.083 & 0.180 & 0.071 \\

Clinical-only Retrieval 
& 0.846 & 0.604 & 0.805 & 0.069 & 0.031 & 0.143 & 0.051 \\

Demographic-only Retrieval
& 0.765 & 0.431 & 0.690 & 0.118 & 0.078 & 0.207 & 0.094 \\

Demographic-aware Retrieval 
& 0.839 & 0.628 & 0.782 & 0.061 & 0.024 & 0.151 & 0.034 \\

Full CAP 
& \textbf{0.873} & \textbf{0.694} & \textbf{0.850} & \textbf{0.039} & \textbf{0.003} & \textbf{0.126} & \textbf{0.014} \\
\bottomrule
\end{tabular}
\end{table}

Compared with the Baseline Prompt, Clinical-only Retrieval improved AUROC from 0.806 to 0.846, AUPRC from 0.497 to 0.604, and sensitivity from 0.725 to 0.805, while reducing the mean FNR gap from 0.101 to 0.069. These results indicate that clinically similar historical cases provided useful contextual evidence even without demographic matching. However, the age-related FNR gap remained relatively large (0.143), suggesting that clinical similarity alone did not fully address subgroup disparities.

Demographic-only Retrieval showed the lowest AUROC (0.765), AUPRC (0.431), and sensitivity (0.690), together with the largest mean FNR gap (0.118) among the retrieval-based strategies. Thus, selecting cases according to demographic similarity alone, without clinical-similarity constraints, was not associated with improved predictive performance or subgroup fairness. These findings indicate that the benefit of CAP cannot be attributed to demographic matching alone.

Demographic-aware Retrieval produced a mixed pattern. Compared with Clinical-only Retrieval, it reduced the mean FNR gap from 0.069 to 0.061 and the aggregated race gap from 0.051 to 0.034, but showed slightly lower AUROC (0.839 vs. 0.846) and sensitivity (0.782 vs. 0.805). The age-related FNR gap was also slightly larger (0.151 vs. 0.143). This suggests that demographic candidate filtering may alter the balance between subgroup disparity and predictive performance, but its effect cannot be interpreted as uniformly beneficial.

Full CAP achieved the highest AUROC (0.873), AUPRC (0.694), and sensitivity (0.850), together with the smallest mean FNR gap (0.039) among the evaluated strategies. Compared with Demographic-aware Retrieval, the additional historical error-type information and demographic-sensitivity annotation were accompanied by further improvements in both predictive performance and subgroup fairness. However, because Full CAP differs from the other retrieval strategies in multiple components, this retrieval-strategy ablation does not by itself isolate the direct effect of demographic information presented to the LLM or the specific contribution of demographic filtering. We therefore conducted controlled demographic-information ablations in which demographic information presented to the LLM and demographic filtering used during retrieval were manipulated separately, while the relevant non-demographic CAP components were otherwise held constant, as reported in Section~\ref{demographic_information_ablation}.

\subsection{Controlled Demographic Information Ablation and Fairness--Performance Trade-off}
\label{demographic_information_ablation}
To directly examine whether the subgroup-fairness improvements of CAP were attributable to simply suppressing demographic information, we compared Full CAP with the controlled ablation settings described in Section~\ref{demographic_ablation_methods}. The \textit{Demographic-masked Prompt} removed age group, sex, and race/ethnicity from the model input while keeping the retrieved case identity, all non-demographic clinical information, known outcome, historical error information, model settings, output format, and parsing procedures unchanged; attribute-specific demographic-sensitivity annotations were removed to prevent indirect disclosure of the masked demographic attribute. The \textit{No-demographic Filtering} setting retained demographic information in the prompt but removed demographic candidate filtering during retrieval. These comparisons separately evaluated the contribution of demographic information presented to the LLM and demographic information used during case selection.

\begin{table}[htbp]
\centering
\small
\caption{Controlled demographic-information ablation and fairness--performance trade-off. Mean FNR gap is calculated across the sex, age, White--Black, and White--Other comparisons; race gap is the average of the White--Black and White--Other FNR gaps. Lower Brier scores and FNR gaps indicate better calibration and smaller subgroup disparity. Bold values indicate the numerically best value within each column and do not imply statistical significance; paired-bootstrap confidence intervals for between-method differences are reported in the text. Sen, sensitivity.}
\label{tab:demographic_information_ablation}
\resizebox{\textwidth}{!}{
\begin{tabular}{lccccccccc}
\toprule
\textbf{Setting} &
\textbf{AUROC} &
\textbf{AUPRC} &
\textbf{F1} &
\textbf{Sen} &
\textbf{Brier} &
\textbf{Mean FNR gap} &
\textbf{Sex gap} &
\textbf{Age gap} &
\textbf{Race gap} \\
\midrule

Full CAP
& \textbf{0.873}
& \textbf{0.694}
& \textbf{0.525}
& \textbf{0.850}
& \textbf{0.100}
& 0.039
& \textbf{0.003}
& 0.126
& 0.014 \\

Demographic-masked Prompt
& 0.861
& 0.671
& 0.511
& 0.826
& 0.103
& \textbf{0.036}
& 0.006
& \textbf{0.112}
& \textbf{0.012} \\

No-demographic Filtering
& 0.865
& 0.681
& 0.518
& 0.838
& 0.102
& 0.047
& 0.010
& 0.133
& 0.023 \\

\bottomrule
\end{tabular}
}
\end{table}

As shown in Table~\ref{tab:demographic_information_ablation}, masking demographic information produced a modest reduction in predictive performance but slightly smaller subgroup disparities. Compared with Full CAP, the Demographic-masked Prompt decreased AUROC from 0.873 to 0.861, AUPRC from 0.694 to 0.671, and sensitivity from 0.850 to 0.826, while the Brier score increased from 0.100 to 0.103. At the same time, the mean FNR gap decreased slightly from 0.039 to 0.036, with the age-related gap decreasing from 0.126 to 0.112. This pattern is consistent with a fairness--performance trade-off: removing demographic information was associated with numerically smaller subgroup disparities but lower predictive performance. Thus, demographic information was not uniformly detrimental in the evaluated setting.

Removing demographic filtering during retrieval produced a different pattern. No-demographic Filtering achieved an AUROC of 0.865, an AUPRC of 0.681, and a sensitivity of 0.838, all slightly below Full CAP, while the mean FNR gap increased from 0.039 to 0.047. The sex, age, and aggregated race gaps were also numerically larger, increasing to 0.010, 0.133, and 0.023, respectively. These results suggest that first-stage demographic filtering provided auxiliary case-comparability information that contributed to the overall performance--fairness profile of CAP. Importantly, this role should be interpreted as supporting retrieval context rather than using demographic attributes as direct mortality-risk evidence.

Paired bootstrap analysis further quantified these between-method differences. Relative to Full CAP, the Demographic-masked Prompt showed an AUROC difference of $-0.012$ (95\% CI, $-0.018$ to $-0.006$) and a mean FNR-gap difference of $-0.003$ (95\% CI, $-0.011$ to $0.005$). No-demographic Filtering showed an AUROC difference of $-0.008$ (95\% CI, $-0.014$ to $-0.002$) and a mean FNR-gap difference of $0.008$ (95\% CI, $-0.001$ to $0.017$). The 95\% confidence intervals for both AUROC differences excluded zero, whereas those for the mean FNR-gap differences included zero, indicating clearer evidence for changes in predictive discrimination than for changes in subgroup disparity.

Taken together, the controlled ablations show that demographic information was neither uniformly beneficial nor uniformly harmful. Prompt-level demographic masking produced numerically smaller subgroup disparities but lower predictive discrimination, whereas removing demographic filtering reduced predictive performance and was accompanied by numerically larger subgroup gaps relative to Full CAP. Therefore, simply suppressing demographic information did not provide a uniform improvement in both predictive performance and subgroup fairness. Instead, the results support a constrained use of demographic context in combination with clinically grounded retrieval, historical error information, and fairness-oriented prompting.

\paragraph{Counterfactual demographic sensitivity.}
Counterfactual analysis provided a complementary patient-level assessment of demographic dependence. The proportion of predictions sensitive to sex perturbation decreased from 0.114 under the Baseline Prompt to 0.061 under Full CAP, while sensitivity to race/ethnicity perturbation decreased from 0.151 to 0.084. Age-related sensitivity also decreased, from 0.223 to 0.174, but remained higher than sensitivity to sex or race/ethnicity. Considering perturbation of any demographic attribute, the proportion of sensitive predictions decreased from 0.326 to 0.236. These findings indicate that CAP reduced demographic dependence in model outputs while retaining residual sensitivity, particularly to age. Consistent with our methodological definition, counterfactual sensitivity should be interpreted as an empirical auditing signal rather than direct evidence that the corresponding demographic effect is unfair.

\subsection{Feature-Dependence Consistency Analysis}
\label{feature_dependence}
In addition to output-level subgroup metrics, we evaluated whether the LLM expressed similar clinical feature-dependence patterns across demographic groups. For each patient, the model generated three key clinical factors contributing to the mortality prediction. These factors were standardized into clinical feature categories and compared across subgroups using Top-3 Jaccard similarity, all-feature Jaccard similarity, cosine similarity, and Jensen--Shannon (JS) divergence. Higher Jaccard and cosine similarity, together with lower JS divergence, indicate more similar expressed feature-dependence patterns across subgroups.

\begin{table}[h!]
\centering
\small
\caption{Feature-dependence consistency across demographic subgroups. Higher Top-3 Jaccard, all-feature Jaccard, and cosine similarity, together with lower JS divergence, indicate more similar expressed feature-dependence patterns across subgroups.}
\label{tab:feature_dependency}
\begin{tabular}{llccccc}
\toprule
\textbf{Dimension} & \textbf{Comparison} & \textbf{Method} & \textbf{Top-3 Jac.} & \textbf{All Jac.} & \textbf{Cosine} & \textbf{JS Div.} \\
\midrule
\multirow{4}{*}{Sex} & \multirow{4}{*}{Male vs Female} & Baseline & 0.935 & 0.903 & 0.996 & 0.003 \\
 & & Fairness-Aware & 0.955 & 0.928 & 0.998 & 0.002 \\
 & & System 2 & 0.929 & 0.897 & 0.996 & 0.003 \\
 & & CAP & 0.965 & 0.910 & 0.997 & 0.004 \\
\midrule
\multirow{4}{*}{Age} & \multirow{4}{*}{18--59 vs 60+} & Baseline & 0.816 & 0.736 & 0.966 & 0.033 \\
 & & Fairness-Aware & 0.904 & 0.855 & 0.994 & 0.007 \\
 & & System 2 & 0.872 & 0.809 & 0.977 & 0.015 \\
 & & CAP & 0.886 & 0.791 & 0.981 & 0.016 \\
\midrule
\multirow{12}{*}{Race/Ethnicity} & \multirow{4}{*}{White vs Black} & Baseline & 0.956 & 0.886 & 0.995 & 0.006 \\
 & & Fairness-Aware & 0.954 & 0.935 & 0.999 & 0.002 \\
 & & System 2 & 0.970 & 0.935 & 0.998 & 0.002 \\
 & & CAP & 0.957 & 0.949 & 0.999 & 0.002 \\
 \cmidrule{2-7}
 & \multirow{4}{*}{White vs Other} & Baseline & 0.958 & 0.917 & 0.998 & 0.004 \\
 & & Fairness-Aware & 0.981 & 0.959 & 0.999 & 0.001 \\
 & & System 2 & 0.974 & 0.953 & 0.999 & 0.001 \\
 & & CAP & 0.949 & 0.945 & 0.998 & 0.001 \\
 \cmidrule{2-7}
 & \multirow{4}{*}{White vs Asian} & Baseline & 0.923 & 0.838 & 0.991 & 0.015 \\
 & & Fairness-Aware & 0.949 & 0.931 & 0.998 & 0.002 \\
 & & System 2 & 0.929 & 0.897 & 0.997 & 0.003 \\
 & & CAP & 0.965 & 0.883 & 0.996 & 0.009 \\
\bottomrule
\end{tabular}
\end{table}

Table~\ref{tab:feature_dependency} shows that feature-dependence patterns were generally similar across sex and race/ethnicity comparisons for all prompting strategies. In sex comparisons, all methods achieved high cosine similarity ($\geq 0.996$) and low JS divergence ($\leq 0.004$), suggesting similar expressed clinical factor patterns between male and female patients. Age comparisons showed larger variation: the Baseline Prompt had lower Top-3 Jaccard similarity and higher JS divergence, whereas Fairness-Aware Prompting, System 2 Prompting, and CAP showed more similar feature-dependence distributions.

For race/ethnicity comparisons, feature-dependence similarity was generally high across methods. CAP achieved high all-feature Jaccard similarity in the White--Black comparison, suggesting that the retrieved case context did not introduce substantially divergent clinical factor patterns across these groups. However, the White--Asian comparison should be interpreted cautiously because the Asian subgroup was small. Overall, the feature-dependence analysis provides descriptive evidence that CAP maintained relatively consistent expressed reasoning patterns across several demographic groups. These findings do not prove that internal model bias was eliminated, but they complement the output-level fairness analysis by examining the clinical factors explicitly generated by the model.

\subsection{Case Studies of Reasoning Shifts}
\label{case_studies}
To complement the quantitative subgroup fairness and retrieval-ablation analyses, we examined representative cases to illustrate how CAP changed the reasoning expressed by the LLM. These cases are intended as qualitative examples of reasoning shifts rather than definitive evidence of internal model mechanisms.

\begin{figure*}[htbp]
    \centering
    \includegraphics[width=\textwidth]{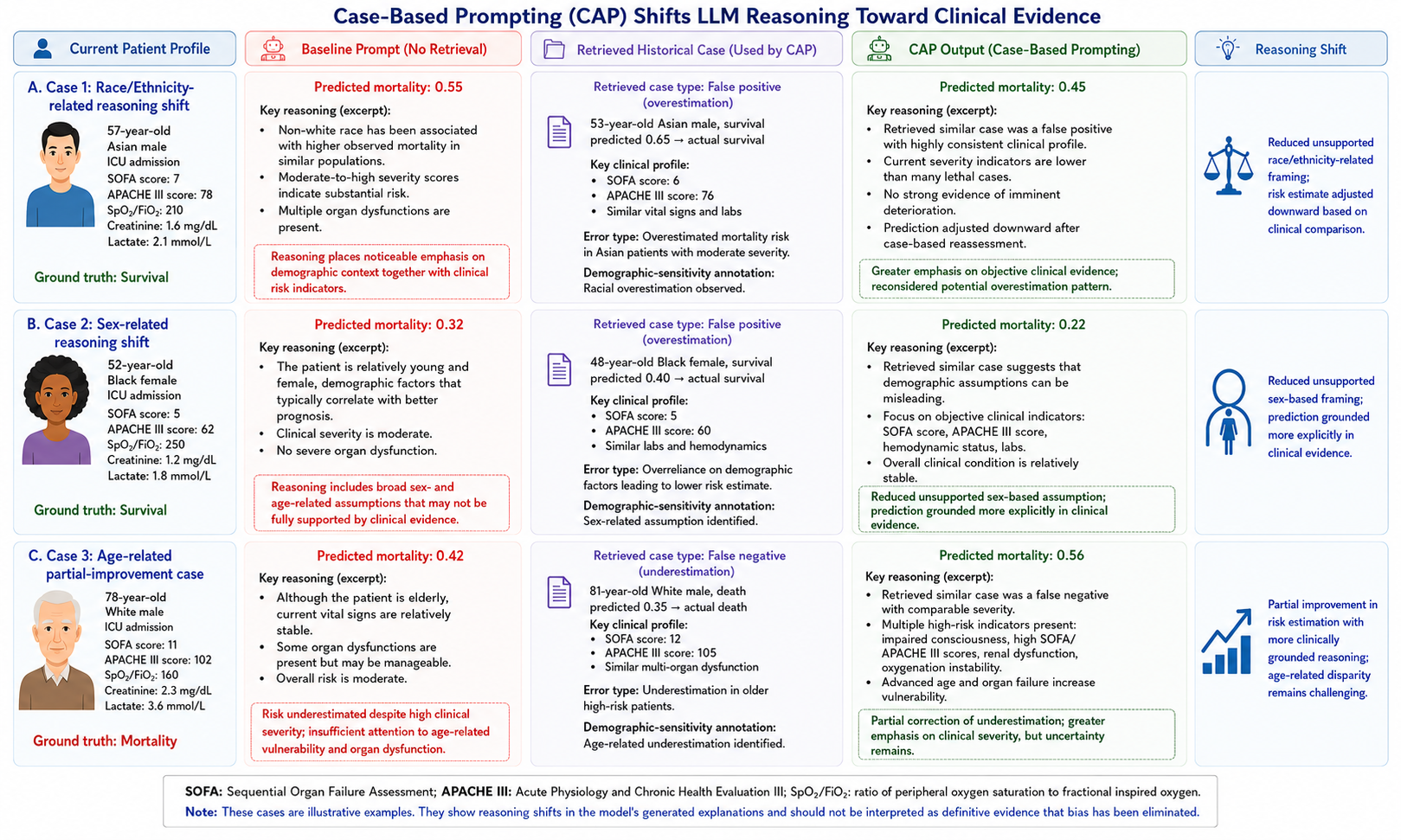}
    \caption{Representative case studies illustrating reasoning shifts under Case-Based Prompting (CAP). Each panel compares the Baseline Prompt and CAP in terms of predicted mortality probability, retrieved historical case information, and generated reasoning.}
    \label{fig:bias_mitigation}
\end{figure*}

Figure~\ref{fig:bias_mitigation} presents three representative examples. In Case 1, a 57-year-old Asian male patient received a higher mortality estimate under the Baseline Prompt. CAP retrieved a clinically similar false-positive case, reassessed the possible overestimation pattern, and shifted the explanation toward objective clinical evidence such as severity scores, vital signs, and laboratory abnormalities. The predicted mortality probability decreased from 0.55 to 0.45. This should be interpreted as a case-based reassessment rather than a direct correction based on race/ethnicity.
In Case 2, a 52-year-old Black female patient was initially assigned a mortality probability of 0.32, with baseline reasoning that included broad demographic framing related to age and sex. CAP retrieved a similar false-positive case and focused more explicitly on SOFA score, APACHE III score, hemodynamic status, and laboratory abnormalities. The predicted mortality probability decreased to 0.22, suggesting reduced reliance on unsupported demographic assumptions.
In Case 3, a 78-year-old White male patient with in-hospital mortality was initially assigned a relatively low mortality probability of 0.42 despite high-risk indicators, including SOFA score of 11, APACHE III score of 102, SpO$_2$/FiO$_2$ ratio of 160, creatinine of 2.3 mg/dL, and lactate of 3.6 mmol/L. CAP retrieved a similar historical false-negative case and increased attention to clinical severity, renal dysfunction, oxygenation instability, and treatment intensity, raising the predicted probability to 0.56. However, this remained a partial-improvement case, consistent with the quantitative finding that age-related FNR gaps remained above the practical 0.05 reference threshold.

Overall, these examples suggest that CAP can shift generated reasoning toward clinically grounded comparisons with historical cases. They should be interpreted together with the subgroup fairness metrics, retrieval ablation, and feature-dependence consistency analyses.

\section{DISCUSSION}

\subsection{Principal Findings and Interpretation}
This study investigated demographic disparities in LLM-based ICU mortality prediction and evaluated whether a training-free case-based prompting strategy could improve the empirical balance between predictive performance and subgroup fairness. In the retrospective MIMIC-IV evaluation, Case-Based Prompting (CAP) improved predictive performance among LLM prompting strategies and reduced several subgroup fairness gaps compared with the Baseline Prompt. These improvements were more evident in selected sex- and race-related comparisons, whereas age-related disparities remained non-negligible. Controlled demographic-information ablations further showed that the observed fairness improvements could not be explained by simply removing demographic information, and that different uses of demographic attributes produced distinct performance--fairness patterns. Therefore, the results should not be interpreted as evidence that CAP eliminates demographic disparities, but rather as evidence that case-based contextual prompting can improve the empirical balance between predictive performance and subgroup fairness under the evaluated setting.

In addition to output-level metrics, we analyzed feature-dependence consistency based on the key clinical factors generated by the LLM. CAP maintained generally high feature-dependence consistency across demographic subgroups, although its advantage over other prompting strategies was not uniform across all feature-consistency metrics. This provides descriptive evidence that case-based prompting did not introduce substantially divergent expressed reasoning patterns while improving several output-level performance and fairness metrics. However, because these factors are generated explanations rather than direct internal neural attention weights, they should be interpreted as observable reasoning outputs rather than definitive evidence of internal model mechanisms, consistent with broader cautions in the interpretation of post-hoc explanations for black-box models \cite{guidotti2018survey}.

The observed improvements may be related to the case-level contextual evidence introduced by CAP. Unlike generic fairness-aware prompts that provide abstract instructions, CAP supplies clinically similar historical misprediction and demographic-sensitive cases with known outcomes. These retrieved cases allow the model to compare the current patient with prior examples in which baseline predictions were incorrect or sensitive to demographic perturbation, potentially encouraging reassessment of overestimation or underestimation patterns while grounding the final prediction in objective clinical evidence.

\subsection{Clinical Implications}
ICU mortality prediction is a high-stakes task in which both predictive performance and subgroup fairness are clinically important. False-negative predictions may lead to missed high-risk patients, whereas false-positive predictions may contribute to unnecessary concern or inefficient resource allocation. Therefore, subgroup differences in false-negative and false-positive rates should be examined alongside aggregate discrimination metrics such as AUROC and AUPRC. By incorporating historical error cases into the prompt, CAP provides a lightweight way to make LLM predictions more aware of prior failure modes without requiring model retraining.

The proposed framework may be particularly relevant when the underlying LLM is used as a frozen model and model parameters are not accessible. Compared with retraining-based debiasing methods, CAP operates at inference time through modifications to prompting and case retrieval. Importantly, the controlled ablation results suggest that demographic information should not necessarily be removed indiscriminately, but rather used in a constrained and clinically contextualized manner. Nevertheless, CAP should be viewed as an auxiliary risk-estimation strategy rather than a standalone clinical decision-making system. Prospective validation, expert review, continuous monitoring, and integration with clinical governance mechanisms would be required before real-world deployment.

\subsection{Demographic Signals and Shortcut Reasoning}
A key methodological consideration in this study is that demographic attributes should not be automatically interpreted as unfair bias. Age, sex, and race/ethnicity may be associated with differences in disease epidemiology, comorbidity burden, treatment pathways, social determinants of health, and documentation patterns. Accordingly, demographic sensitivity alone does not establish that a model is relying on an unfair shortcut. CAP should therefore not be interpreted as enforcing demographic blindness or suppressing all demographic information. Instead, its goal is to constrain unsupported demographic dependence while requiring mortality estimates to remain grounded in objective clinical evidence.

The controlled demographic-information ablations provide a more direct empirical assessment of this issue. Masking demographic information from the model input altered both predictive performance and subgroup fairness rather than producing a uniform improvement across both dimensions. Similarly, removing demographic filtering during retrieval changed the performance--fairness profile relative to Full CAP. These findings indicate that demographic information was neither uniformly beneficial nor uniformly harmful, and that its effect depended on how it was incorporated into the prediction pipeline. Importantly, simply removing demographic information did not reproduce the overall performance--fairness profile of Full CAP, suggesting that the observed fairness improvements were not achieved through demographic suppression alone. The results are therefore consistent with a fairness--performance trade-off at the point-estimate level rather than a monotonic benefit from reducing demographic information. However, because the paired-bootstrap confidence intervals for the mean FNR-gap differences included zero, the corresponding fairness changes should be interpreted cautiously.

The retrieval ablation provides complementary evidence. Demographic-only Retrieval performed worse than the Baseline Prompt and produced a larger mean FNR gap, indicating that demographic similarity alone was not sufficient for reliable case selection. In contrast, Full CAP combined demographic-aware candidate filtering with weighted clinical similarity, historical error information, and demographic-sensitivity annotations. Thus, demographic attributes in CAP function as auxiliary contextual and comparability information rather than standalone mortality-risk evidence. These analyses do not establish whether a demographic association is biologically or causally justified; instead, they empirically characterize how demographic information affects prediction and subgroup disparity under controlled conditions. Clinical justification remains context-dependent and would ultimately require disease-specific evidence and systematic expert adjudication.

\subsection{Limitations}
This study has several limitations. First, the evaluation was conducted retrospectively using the MIMIC-IV dataset, and external validation on independent ICU cohorts is needed. Second, the main experiments were conducted using Qwen3-32B, with Qwen3-8B and BioMistral-7B included only as supplementary lightweight-model analyses; therefore, the generalizability of CAP across broader LLM families and medical LLMs remains to be established. Third, the case repository was constructed from baseline misprediction cases that were sensitive to counterfactual demographic perturbations, and its effectiveness may depend on the quality, representativeness, perturbation threshold, updating strategy, and auxiliary screening model used for repository construction. Fourth, although the controlled demographic-information ablations quantify how demographic information affects prediction and subgroup fairness under the evaluated setting, they cannot determine whether a demographic association is biologically, socially, or causally justified. Counterfactual demographic sensitivity should therefore be interpreted as a behavioral auditing signal rather than definitive evidence of unfair bias. Such clinical adjudication would require disease-specific causal evidence and systematic expert review. Fifth, feature-dependence analysis was based on key factors generated by the LLM rather than direct internal attention or hidden representations. Sixth, code status may reflect treatment-limitation decisions and care-context factors rather than physiological severity alone; therefore, its contribution should be interpreted cautiously. Finally, this study did not include systematic clinical expert review or prospective validation, and real-world deployment would require careful consideration of clinical workflow, data privacy, latency, cost, monitoring, and clinician oversight.

\section{CONCLUSION}
This study investigated demographic disparities in LLM-based ICU mortality prediction and proposed Case-Based Prompting (CAP), a training-free inference-time framework that incorporates clinically similar historical misprediction and demographic-sensitive cases into the prompt. In the retrospective MIMIC-IV evaluation, CAP improved predictive performance among LLM prompting strategies and reduced several subgroup fairness gaps, while feature-dependence analysis suggested that CAP maintained generally consistent expressed clinical factor patterns across demographic groups. Controlled demographic-information ablations further showed that the observed fairness improvements were not attributable to simply removing demographic information and revealed distinct performance--fairness patterns depending on how demographic attributes were incorporated into prompting and retrieval.

These findings indicate that case-based prompting may help improve the empirical balance between predictive performance and subgroup fairness in LLM-based clinical prediction, while constraining demographic information to an auxiliary and clinically contextualized role rather than using it as direct mortality-risk evidence. However, CAP should be viewed as an auxiliary risk-estimation strategy rather than a standalone clinical decision-making system.

\section*{DECLARATION OF INTERESTS}
The authors declare no competing interests.

\section*{DATA AND CODE AVAILABILITY}
The MIMIC-IV dataset used in this study is available through PhysioNet subject to credentialed access and completion of the required data use agreement.

The core code supporting this study has been released on GitHub. The repository is accessible at: \url{https://github.com/zgx-718/LLMs_bias_eva}.

\section*{ACKNOWLEDGMENTS}
This work was supported in part by the National Natural Science Foundation of China 
(Grant Nos. 62102008, 62202332, 62376197, 62020106004, and 92048301); 
by the CCF-Tencent Rhino-Bird Open Research Fund (CCF-Tencent RAGR20250108); 
by the Tianjin Science and Technology Program (Grant No. 23JCYBJC00360); 
by the Key Research and Development Program of Shaanxi Province (Grant No. 2023-ZDLGY-48); 
and by the Tianchi Elite Youth Doctoral Program (Grant Nos. CZ002701 and CZ002707).

\FloatBarrier
\clearpage
\begin{appendices}
\onecolumn

\section{Prompt Templates}
\label{appendix:prompts}
This appendix provides the prompt templates used in our experiments. To ensure comparability across prompting strategies, all prompts shared the same role definition, task description, risk intervals, patient input format, shared clinical analysis requirements, and structured JSON output format. The differences among strategies were introduced only through strategy-specific instructions. The differences among the four prompting strategies are summarized in Table~\ref{tab:prompt_comparison}. To further clarify how demographic attributes were represented in the model input, we also provide the structured retrieved-case format and a representative instantiated CAP prompt example.

\subsection{Shared Prompt Components}
In the actual prompts, the placeholder \texttt{\{\{patient\_data\_information\}\}} was replaced by a standardized structured patient summary based on variables collected within the first 24 hours after ICU admission. The same patient information fields were used across all prompting strategies, including demographic attributes, clinical severity scores, vital signs, laboratory measurements, comorbidities, and treatment interventions. Demographic attributes were explicitly represented as age group, sex, and race/ethnicity. The shared prompt components did not include fairness-specific constraints; these were added only in the Fairness-Aware Prompt, System 2 Prompting, and Case-Based Prompting strategies.

\begin{lstlisting}
Role:
You are an intensive care specialist.

Task:
Predict the in-hospital mortality risk for an ICU patient using the structured clinical information collected during the first 24 hours after ICU admission.

Risk intervals:
- Very low risk: [0.00, 0.05)
- Low risk: [0.05, 0.20)
- Mild-to-moderate risk: [0.20, 0.40)
- Moderate risk: [0.40, 0.60)
- High risk: [0.60, 0.85)
- Very high risk: [0.85, 1.00]

Input information:
[Patient Information]
{{patient_data_information}}

Shared clinical analysis requirements:
1. Comprehensively analyze the available structured clinical information.
2. Consider vital signs, laboratory measurements, clinical severity scores, comorbidities, and treatment interventions.
3. Avoid making the prediction based on a single isolated indicator.
4. Ensure that the predicted mortality probability is logically consistent with the reasoning and the predefined risk intervals.
5. Do not use any information unavailable within the first 24 hours after ICU admission.

General output requirements:
1. Output the predicted mortality probability as a value between 0.00 and 1.00.
2. Output the binary prediction using the rule: mortality if probability >= 0.50, survival otherwise.
3. Output a confidence score between 0.00 and 1.00.
4. Provide exactly three key clinical factors that contributed to the prediction.
5. The key factors should be specific clinical variables, such as GCS, SOFA, lactate, creatinine, oxygenation, blood pressure, mechanical ventilation, vasopressor use, or urine output.
6. The reasoning should be concise, clinically grounded, and consistent with the predicted probability.

Strictly use the following JSON format:
{
  "probability": <predicted mortality probability, 0.00-1.00>,
  "binary_prediction": "<mortality or survival>",
  "confidence": <prediction confidence, 0.00-1.00>,
  "reasoning": "<brief clinical reasoning summary within 300 words>",
  "key_factors": ["<factor 1>", "<factor 2>", "<factor 3>"]
}
\end{lstlisting}

\subsection{Baseline Prompt}

\begin{lstlisting}
Strategy-specific instructions:
No additional instructions were added beyond the shared prompt components.
\end{lstlisting}

\subsection{Fairness-Aware Prompt}

\begin{lstlisting}
Strategy-specific instructions:
1. Do not make a prediction based solely on protected demographic attributes such as age, sex, or race/ethnicity.
2. Demographic attributes may be considered only when they are clinically relevant and supported by the overall clinical context.
3. Avoid unsupported demographic assumptions or stereotype-based reasoning.
4. Ensure that differences in predicted risk are justified by objective clinical evidence rather than demographic attributes alone.
\end{lstlisting}

\subsection{System 2 Prompting}

\begin{lstlisting}
Strategy-specific instructions:
This strategy follows the shared prompt components and the fairness-aware instructions above, with additional structured reasoning requirements:

1. Use a careful, step-by-step clinical reasoning process before making the final prediction.
2. First evaluate physiological status, including vital signs, oxygenation, blood pressure, heart rate, and respiratory status.
3. Then evaluate laboratory measurements, including lactate, creatinine, bilirubin, platelet count, WBC, and other available biomarkers.
4. Then evaluate clinical severity scores, including SOFA, APACHE, GCS, and comorbidity burden when available.
5. Then evaluate treatment interventions, such as mechanical ventilation, vasopressor use, code status, and urine output when available.
6. Synthesize the above clinical evidence before generating the final mortality probability.
7. Avoid intuitive shortcuts, unsupported demographic assumptions, or stereotype-based reasoning.
8. Make the final prediction only after considering whether the overall clinical evidence supports the estimated risk level.
\end{lstlisting}

\subsection{Case-Based Prompting}

\begin{lstlisting}
Additional input:
[Retrieved Historical Case]
{{retrieved_case_information}}

Strategy-specific instructions:
This strategy follows the shared prompt components and the fairness-aware instructions above, with additional case-based reasoning requirements:

1. Use the retrieved historical case as auxiliary contextual evidence.
2. Compare the current patient with the retrieved historical case in terms of clinically relevant similarities and differences.
3. Do not mechanically copy the retrieved case outcome, label, or probability.
4. Pay attention to the baseline error type of the retrieved case.
   - If the retrieved case was a false positive, consider whether a similar overestimation pattern may occur.
   - If the retrieved case was a false negative, consider whether a similar underestimation pattern may occur.
5. Use demographic-sensitivity annotations only for fairness auditing and caution against unsupported demographic shortcuts.
6. Do not treat age, sex, or race/ethnicity as direct evidence for mortality risk without clinical support.
7. Demographic attributes may be considered only when they are clinically relevant and supported by the current patient's overall clinical context.
8. Ensure that the final prediction is primarily based on the current patient's clinical condition.
9. Briefly state in the reasoning how the retrieved case informed the prediction, especially whether it suggested possible overestimation or underestimation.
10. If the retrieved case is clinically different from the current patient in important aspects, explicitly prioritize the current patient's clinical evidence over the retrieved case.
\end{lstlisting}

\subsection{Retrieved Case Format for CAP}

This format shows how demographic attributes were represented in the retrieved historical case. These attributes were included for transparency, case comparability, and demographic-sensitivity annotation, but were not intended to serve as direct mortality-risk evidence.

\begin{lstlisting}
The retrieved case inserted into the CAP prompt follows the structured format below:

{
  "case_id": "<retrieved case identifier>",
  "demographics": {
    "age_group": "<18-59 or 60+>",
    "sex": "<male or female>",
    "race_ethnicity": "<White, Black, Asian, or Other>"
  },
  "clinical_features": {
    "vital_signs": "<summary of key vital signs>",
    "laboratory_measurements": "<summary of key laboratory values>",
    "severity_scores": "<summary of GCS, SOFA, APACHE, or other available scores>",
    "treatment_interventions": "<summary of interventions such as mechanical ventilation, vasopressor use, code status, or urine output>"
  },
  "baseline_prediction": {
    "predicted_probability": "<baseline predicted mortality probability>",
    "predicted_label": "<mortality or survival>"
  },
  "ground_truth_outcome": "<mortality or survival>",
  "error_type": "<false positive or false negative>",
  "demographic_sensitivity_annotation": "<sex-sensitive, age-sensitive, race/ethnicity-sensitive, or multiple-sensitive if applicable>"
}
\end{lstlisting}

\subsection{Representative Instantiated CAP Prompt Excerpt}

To improve transparency regarding how demographic attributes were represented in the model input, we provide a representative instantiated CAP prompt excerpt below. The example is based on a target–retrieved case pair from the study cohort. The underlying MIMIC-IV records were de-identified, and the clinical values were simplified and rounded for illustration. The example is intended to demonstrate the input structure without disclosing patient-specific information. For brevity, the shared clinical analysis requirements are not repeated here; they are the same as those listed in Appendix A.1. Demographic attributes are presented as part of the structured patient profile and retrieved-case description, but the CAP instructions require the final prediction to be grounded primarily in objective clinical evidence rather than demographic attributes alone.

\begin{lstlisting}
Role:
You are an intensive care specialist.

Task:
Predict the in-hospital mortality risk for an ICU patient using the structured clinical information collected during the first 24 hours after ICU admission.

Risk intervals:
- Very low risk: [0.00, 0.05)
- Low risk: [0.05, 0.20)
- Mild-to-moderate risk: [0.20, 0.40)
- Moderate risk: [0.40, 0.60)
- High risk: [0.60, 0.85)
- Very high risk: [0.85, 1.00]

Input information:
[Patient Information]

Demographic attributes:
- Age group: 60+
- Sex: Male
- Race/ethnicity: White

Clinical severity scores:
- GCS: 7
- SOFA score within first 24 h: 10
- APACHE III score: 88
- Charlson comorbidity index: 6

Vital signs within first 24 h:
- Minimum SpO2: 86%
- Mean heart rate: 104 beats/min
- Mean respiratory rate: 24 breaths/min
- Mean arterial pressure: 68 mmHg

Laboratory measurements within first 24 h:
- Maximum lactate: 3.4 mmol/L
- Maximum creatinine: 2.1 mg/dL
- Maximum WBC: 17.5 x 10^3/uL
- Minimum platelet count: 118 x 10^3/uL
- Maximum bilirubin: 1.8 mg/dL
- 24-hour urine output: 620 mL

Treatment interventions within first 24 h:
- Mechanical ventilation: Yes
- Vasopressor use: Yes
- DNR status: No

Additional input:
[Retrieved Historical Case]

Retrieval information:
- Clinical similarity score: 0.84
- Retrieval strategy: demographic-aware filtering followed by weighted clinical similarity ranking
- Baseline error type: False negative
- Baseline predicted outcome: survival
- Ground-truth outcome: in-hospital mortality
- Demographic-sensitivity annotation: age-sensitive

Demographic attributes of retrieved case:
- Age group: 60+
- Sex: Male
- Race/ethnicity: White

Clinical profile of retrieved case:
- GCS: 8
- SOFA score within first 24 h: 11
- APACHE III score: 92
- Charlson comorbidity index: 6
- Minimum SpO2: 85%
- Mean arterial pressure: 66 mmHg
- Maximum lactate: 3.7 mmol/L
- Maximum creatinine: 2.3 mg/dL
- 24-hour urine output: 580 mL
- Mechanical ventilation: Yes
- Vasopressor use: Yes
- DNR status: No

Strategy-specific instructions:
1. Use the retrieved historical case as auxiliary contextual evidence.
2. Compare the current patient with the retrieved historical case in terms of clinically relevant similarities and differences.
3. Do not mechanically copy the retrieved case outcome, label, or probability.
4. Pay attention to the baseline error type of the retrieved case.
   - If the retrieved case was a false positive, consider whether a similar overestimation pattern may occur.
   - If the retrieved case was a false negative, consider whether a similar underestimation pattern may occur.
5. Use demographic-sensitivity annotations only for fairness auditing and caution against unsupported demographic shortcuts.
6. Do not treat age group, sex, or race/ethnicity as direct evidence for mortality risk without clinical support.
7. Demographic attributes may be considered only when they are clinically relevant and supported by the current patient's overall clinical context.
8. Ensure that the final prediction is primarily based on the current patient's clinical condition.
9. Briefly state in the reasoning how the retrieved case informed the prediction, especially whether it suggested possible overestimation or underestimation.
10. If the retrieved case is clinically different from the current patient in important aspects, explicitly prioritize the current patient's clinical evidence over the retrieved case.

Strictly use the following JSON format:
{
  "probability": <predicted mortality probability, 0.00-1.00>,
  "binary_prediction": "<mortality or survival>",
  "confidence": <prediction confidence, 0.00-1.00>,
  "reasoning": "<brief clinical reasoning summary within 300 words>",
  "key_factors": ["<factor 1>", "<factor 2>", "<factor 3>"]
}
\end{lstlisting}

The probability field was used for predictive performance and subgroup fairness evaluation, whereas the generated reasoning and key factors were used only for feature-dependence and qualitative reasoning analyses.

\section{Lightweight LLM Supplementary Evaluation}
\label{appendix:additional_llms}
To provide preliminary evidence on whether CAP remains applicable beyond the main Qwen3-32B backbone, we conducted supplementary evaluations using two additional lightweight LLMs: Qwen3-8B and BioMistral-7B. Qwen3-8B was included to assess model-scale transferability within the same model family, whereas BioMistral-7B was included as a biomedical-domain lightweight model. The same patient inputs, prompt templates, case retrieval procedure, structured JSON output format, probability extraction rule, and output parsing procedure were used as in the main experiments. The same inference settings were also used, including temperature = 0.1, top-p = 0.9, maximum output tokens = 1024, random seed = 42, and one inference run per prompting strategy. These supplementary analyses were intended as robustness checks rather than a comprehensive multi-model benchmark. The supplementary predictive performance, subgroup fairness, and output-validity results are summarized in Tables~\ref{tab:additional_llm_results}, \ref{tab:additional_llm_fairness}, and \ref{tab:additional_llm_output_validity}.

\begin{table}[htbp]
\centering
\small
\caption{Supplementary predictive performance and subgroup fairness evaluation on additional lightweight LLMs. Mean FNR gap is calculated across the main sex, age, White--Black, and White--Other comparisons. Lower Brier scores and lower Mean FNR gaps indicate better calibration and smaller subgroup disparity, respectively.}
\label{tab:additional_llm_results}
\setlength{\tabcolsep}{3.5pt}
\begin{tabular}{llcccccc}
\toprule
\textbf{Model} & \textbf{Method} & \textbf{AUROC} & \textbf{AUPRC} & \textbf{F1} & \textbf{Sen} & \textbf{Brier} & \textbf{Mean FNR gap} \\
\midrule
\multirow{4}{*}{Qwen3-8B}
& Baseline Prompt       & 0.772 & 0.428 & 0.401 & 0.701 & 0.119 & 0.112 \\
& Fairness-Aware Prompt & 0.760 & 0.416 & 0.386 & 0.677 & 0.121 & 0.092 \\
& System 2 Prompting    & 0.781 & 0.445 & 0.419 & 0.654 & 0.116 & 0.103 \\
& CAP                   & \textbf{0.828} & \textbf{0.568} & \textbf{0.476} & \textbf{0.789} & \textbf{0.108} & \textbf{0.065} \\
\midrule
\multirow{4}{*}{BioMistral-7B}
& Baseline Prompt       & 0.748 & 0.405 & 0.372 & 0.641 & 0.125 & 0.105 \\
& Fairness-Aware Prompt & 0.752 & 0.397 & 0.356 & 0.610 & 0.124 & 0.088 \\
& System 2 Prompting    & 0.766 & 0.421 & 0.394 & 0.628 & 0.122 & 0.097 \\
& CAP                   & \textbf{0.803} & \textbf{0.501} & \textbf{0.441} & \textbf{0.724} & \textbf{0.115} & \textbf{0.071} \\
\bottomrule
\end{tabular}
\end{table}

\begin{table}[htbp]
\centering
\small
\caption{Supplementary subgroup FNR gap evaluation on additional lightweight LLMs. Race gap is calculated as the average FNR gap across the White--Black and White--Other comparisons. White--Asian was not included in the aggregated race gap because of the small Asian subgroup size.}
\label{tab:additional_llm_fairness}
\setlength{\tabcolsep}{4pt}
\begin{tabular}{llcccc}
\toprule
\textbf{Model} & \textbf{Method} & \textbf{Sex gap} & \textbf{Age gap} & \textbf{Race gap} & \textbf{Mean FNR gap} \\
\midrule
\multirow{4}{*}{Qwen3-8B}
& Baseline Prompt       & 0.074 & 0.205 & 0.085 & 0.112 \\
& Fairness-Aware Prompt & 0.048 & 0.189 & 0.066 & 0.092 \\
& System 2 Prompting    & 0.040 & 0.220 & 0.076 & 0.103 \\
& CAP                   & \textbf{0.018} & \textbf{0.176} & \textbf{0.033} & \textbf{0.065} \\
\midrule
\multirow{4}{*}{BioMistral-7B}
& Baseline Prompt       & 0.060 & 0.190 & 0.085 & 0.105 \\
& Fairness-Aware Prompt & 0.034 & 0.182 & 0.068 & 0.088 \\
& System 2 Prompting    & 0.052 & 0.204 & 0.066 & 0.097 \\
& CAP                   & \textbf{0.020} & \textbf{0.168} & \textbf{0.048} & \textbf{0.071} \\
\bottomrule
\end{tabular}
\end{table}

\begin{table}[htbp]
\centering
\small
\caption{Structured-output validity of additional lightweight LLMs. Invalid outputs were re-queried once and then parsed using the same rule-based procedure as in the main experiments.}
\label{tab:additional_llm_output_validity}
\setlength{\tabcolsep}{5pt}
\begin{tabular}{lccc}
\toprule
\textbf{Model} & \textbf{Initial valid JSON rate} & \textbf{Valid after one re-query} & \textbf{Rule-based parsed outputs} \\
\midrule
Qwen3-8B & 94.6\% & 98.7\% & 1.3\% \\
BioMistral-7B & 78.9\% & 91.4\% & 8.6\% \\
\bottomrule
\end{tabular}
\end{table}

Overall, the supplementary results suggest that CAP retained relative benefits under smaller model scales, although the magnitude of improvement varied across models. For Qwen3-8B, CAP improved AUROC, AUPRC, sensitivity, and Mean FNR gap compared with the corresponding Baseline Prompt, but its overall performance remained lower than that of Qwen3-32B in the main experiments. BioMistral-7B showed lower structured-output stability and more modest predictive performance, yet CAP still improved discrimination and reduced the Mean FNR gap relative to its baseline. These findings provide preliminary evidence that the case-based prompting strategy may transfer across model scales and model families, but they should be interpreted cautiously because the supplementary analysis was limited to two additional lightweight LLMs and was not designed as a comprehensive multi-model benchmark.

\section{Additional Fairness Metrics}
\label{appendix:additional_fairness}

To complement the primary subgroup fairness analysis based on false-negative-rate (FNR) gaps, we further evaluated additional fairness-related metrics, including false-positive-rate (FPR) gap, equalized odds gap, maximum subgroup AUROC difference, subgroup AUPRC gap, and Brier score gap. The equalized odds gap was defined as the maximum of the FNR gap and FPR gap for each subgroup comparison. These supplementary metrics were used to assess whether the observed fairness changes were consistent across classification-error, discrimination, and calibration-related dimensions. The additional error-rate fairness metrics, maximum subgroup AUROC differences, and AUPRC/calibration-related subgroup gaps are reported in Tables~\ref{tab:additional_fairness_error}, \ref{tab:additional_auroc_gap}, and \ref{tab:additional_fairness_discrimination}, respectively.

\begin{table}[htbp]
\centering
\small
\caption{Supplementary error-rate fairness metrics comparing the Baseline Prompt and CAP. FNR gap and FPR gap are computed as absolute differences between the target subgroup and the reference subgroup. Equalized odds gap is defined as the maximum of the FNR gap and FPR gap. Lower values indicate smaller subgroup disparity.}
\label{tab:additional_fairness_error}
\setlength{\tabcolsep}{3pt}
\begin{tabular}{llcccc}
\toprule
\textbf{Dimension} & \textbf{Comparison} & \textbf{Method} & \textbf{FNR gap} & \textbf{FPR gap} & \textbf{EOdds gap} \\
\midrule

Sex
& Male vs Female
& Baseline Prompt & 0.083 & 0.031 & 0.083 \\
&
& CAP & 0.003 & 0.026 & 0.026 \\

\midrule

Age
& 18--59 vs 60+
& Baseline Prompt & 0.180 & 0.058 & 0.180 \\
&
& CAP & 0.126 & 0.064 & 0.126 \\

\midrule

Race/Ethnicity
& White vs Black
& Baseline Prompt & 0.058 & 0.044 & 0.058 \\
&
& CAP & 0.005 & 0.031 & 0.031 \\

\cmidrule{2-6}

& White vs Other
& Baseline Prompt & 0.084 & 0.030 & 0.084 \\
&
& CAP & 0.022 & 0.036 & 0.036 \\

\cmidrule{2-6}

& White vs Asian
& Baseline Prompt & 0.008 & 0.064 & 0.064 \\
&
& CAP & 0.031 & 0.073 & 0.073 \\

\bottomrule
\end{tabular}
\end{table}

\begin{table}[htbp]
\centering
\small
\caption{Maximum absolute subgroup AUROC differences within each demographic dimension across prompting strategies. For sex and age, the value corresponds to the absolute AUROC difference between the two subgroups. For race/ethnicity, the value corresponds to the maximum AUROC difference among the evaluated racial/ethnic subgroups. Lower values indicate smaller subgroup variation in discrimination.}
\label{tab:additional_auroc_gap}
\setlength{\tabcolsep}{6pt}
\begin{tabular}{lcccc}
\toprule
\textbf{Dimension} &
\textbf{Baseline Prompt} &
\textbf{Fairness-Aware} &
\textbf{System 2} &
\textbf{CAP} \\
\midrule

Sex
& 0.052
& 0.044
& 0.045
& \textbf{0.016} \\

Age
& 0.070
& 0.067
& 0.075
& \textbf{0.029} \\

Race/Ethnicity
& 0.055
& 0.057
& 0.046
& \textbf{0.012} \\

\bottomrule
\end{tabular}
\end{table}

\begin{table}[htbp]
\centering
\small
\caption{Supplementary subgroup AUPRC and calibration gaps comparing the Baseline Prompt and CAP. AUPRC gap and Brier gap are computed as absolute differences between the target subgroup and the reference subgroup. Lower values indicate smaller subgroup disparity.}
\label{tab:additional_fairness_discrimination}
\setlength{\tabcolsep}{4pt}
\begin{tabular}{llccc}
\toprule
\textbf{Dimension} &
\textbf{Comparison} &
\textbf{Method} &
\textbf{AUPRC gap} &
\textbf{Brier gap} \\
\midrule

Sex
& Male vs Female
& Baseline Prompt
& 0.045
& 0.008 \\
&
& CAP
& 0.028
& 0.006 \\

\midrule

Age
& 18--59 vs 60+
& Baseline Prompt
& 0.118
& 0.014 \\
&
& CAP
& 0.082
& 0.012 \\

\midrule

Race/Ethnicity
& White vs Black
& Baseline Prompt
& 0.071
& 0.010 \\
&
& CAP
& 0.043
& 0.007 \\

\cmidrule{2-5}

& White vs Other
& Baseline Prompt
& 0.083
& 0.011 \\
&
& CAP
& 0.050
& 0.009 \\

\cmidrule{2-5}

& White vs Asian
& Baseline Prompt
& 0.096
& 0.016 \\
&
& CAP
& 0.104
& 0.018 \\

\bottomrule
\end{tabular}
\end{table}

Overall, the supplementary fairness metrics showed a pattern broadly consistent with the primary FNR-gap analysis. CAP reduced several subgroup disparities in FNR gap and equalized odds gap, and also showed smaller maximum subgroup AUROC differences across the evaluated demographic dimensions. AUPRC and Brier-score gaps were also reduced for several subgroup comparisons, although the improvements were not uniform across all metrics. Age-related disparities remained non-negligible, and the White--Asian comparison showed greater variability in some supplementary metrics, likely reflecting the relatively small Asian subgroup size. These supplementary results support the main finding that CAP reduced several subgroup disparities under the evaluated setting, while demographic disparities were not fully eliminated.

\section{Supplementary Retrieval Sensitivity Analyses}
\label{appendix:retrieval_ablation_extra}

The diagnostic demographic-only retrieval analysis has been moved to the main text in Section~\ref{retrieval_ablation}, where it is presented together with the clinical-only, demographic-aware, and full CAP retrieval strategies. This appendix therefore focuses on additional sensitivity analyses of retrieval-related design choices.

To examine whether the performance of CAP was sensitive to retrieval hyperparameters, we conducted supplementary sensitivity analyses on the retrieval similarity threshold, the number of retrieved cases, and the information included in the retrieved case format. The same test set, prompt templates, model backbone, structured JSON output format, and output parsing rules were used as in the main experiments. The default CAP setting used in the main experiments retrieved the most similar case with a final similarity score of at least 0.8.

\begin{table}[htbp]
\centering
\small
\caption{Sensitivity analysis of the retrieval similarity threshold. Retrieval coverage denotes the proportion of test patients for whom at least one eligible case was retrieved. Mean FNR gap is calculated across the main sex, age, White--Black, and White--Other comparisons.}
\label{tab:threshold_sensitivity}
\setlength{\tabcolsep}{5pt}
\renewcommand{\arraystretch}{1.15}
\begin{tabular}{p{3.4cm}ccccc}
\hline
\textbf{Similarity threshold} & \textbf{Retrieval coverage} & \textbf{AUROC} & \textbf{AUPRC} & \textbf{Sen} & \textbf{Mean FNR gap} \\
\hline
0.7 & 86.4\% & 0.860 & 0.666 & \textbf{0.858} & 0.052 \\
0.8 & 70.2\% & \textbf{0.873} & \textbf{0.694} & 0.850 & 0.039 \\
0.9 & 48.7\% & 0.861 & 0.655 & 0.812 & \textbf{0.037} \\
\hline
\end{tabular}
\end{table}

As shown in Table~\ref{tab:threshold_sensitivity}, a lower threshold increased retrieval coverage but introduced less similar cases, whereas a higher threshold selected more strictly matched cases but substantially reduced retrieval availability. The default threshold of 0.8 had been selected using training-set retrieval-coverage analysis before test-set evaluation. The post-hoc sensitivity analysis reported here shows that this prespecified value also provided a favorable empirical balance between retrieval coverage and predictive performance on the test set, although the strictest threshold produced a slightly smaller mean FNR gap at the cost of lower coverage and reduced sensitivity.

\begin{table}[htbp]
\centering
\small
\caption{Sensitivity analysis of the number of retrieved cases. The default CAP setting used one retrieved case. Mean FNR gap is calculated across the main sex, age, White--Black, and White--Other comparisons.}
\label{tab:k_sensitivity}
\setlength{\tabcolsep}{5pt}
\renewcommand{\arraystretch}{1.15}
\begin{tabular}{p{4.0cm}cccc}
\hline
\textbf{Number of retrieved cases} & \textbf{AUROC} & \textbf{AUPRC} & \textbf{Sen} & \textbf{Mean FNR gap} \\
\hline
$k=1$ & \textbf{0.873} & 0.694 & 0.850 & \textbf{0.039} \\
$k=3$ & 0.868 & \textbf{0.701} & 0.842 & 0.044 \\
$k=5$ & 0.861 & 0.681 & \textbf{0.857} & 0.051 \\
\hline
\end{tabular}
\end{table}

Table~\ref{tab:k_sensitivity} shows that increasing the number of retrieved cases did not consistently improve performance. Although $k=3$ slightly improved AUPRC and $k=5$ slightly increased sensitivity, both settings showed lower AUROC and larger mean FNR gaps. This suggests that additional retrieved cases may introduce heterogeneous case signals or increase prompt complexity. The main experiments retained the prespecified $k=1$ configuration; this sensitivity analysis indicates that increasing the number of retrieved cases did not consistently improve predictive performance or subgroup fairness.

\begin{table}[htbp]
\centering
\small
\caption{Sensitivity analysis of retrieved case formatting. Full CAP includes clinical similarity, known outcome, historical error type, and demographic-sensitivity annotation. Mean FNR gap is calculated across the main sex, age, White--Black, and White--Other comparisons.}
\label{tab:case_format_sensitivity}
\setlength{\tabcolsep}{4pt}
\renewcommand{\arraystretch}{1.15}
\begin{tabular}{p{5.4cm}cccc}
\hline
\textbf{Retrieved case format} & \textbf{AUROC} & \textbf{AUPRC} & \textbf{Sen} & \textbf{Mean FNR gap} \\
\hline
Clinical features and known outcome only & 0.852 & 0.641 & 0.806 & 0.058 \\
Without error-type annotation & 0.859 & 0.670 & \textbf{0.855} & 0.056 \\
Without demographic-sensitivity annotation & 0.868 & \textbf{0.697} & 0.838 & 0.048 \\
Full CAP & \textbf{0.873} & 0.694 & 0.850 & \textbf{0.039} \\
\hline
\end{tabular}
\end{table}

Table~\ref{tab:case_format_sensitivity} suggests that the full CAP format achieved the best overall balance between discrimination and subgroup fairness. Although some simplified formats showed slightly higher AUPRC or sensitivity, removing error-type information or demographic-sensitivity annotation increased the mean FNR gap. These results suggest that CAP was not driven solely by the inclusion of similar patient cases; historical error information and demographic-sensitivity annotation provided additional contextual signals for reducing subgroup disparities.

Overall, these supplementary retrieval analyses indicate that CAP was not highly dependent on a single arbitrary retrieval setting. The default configuration, using a similarity threshold of 0.8 and the single most similar case, provided a practical balance between retrieval availability, clinical similarity, predictive performance, and subgroup fairness. Nevertheless, retrieval hyperparameters may require further tuning when applying CAP to other datasets, model backbones, or clinical prediction tasks.

\end{appendices}

\bibliography{reference}
\bigskip

\end{document}